\documentclass[10pt,twocolumn,letterpaper]{article}

\usepackage[pagenumbers]{wacv} 

\definecolor{wacvblue}{rgb}{0.21,0.49,0.74}
\usepackage[pagebackref,breaklinks,colorlinks,allcolors=wacvblue]{hyperref}

\usepackage{acronym}
\usepackage{wrapfig}
\usepackage{adjustbox}
\usepackage{makecell}
\usepackage{afterpage}
\usepackage{mathtools}
\usepackage{stfloats}
\usepackage{svg}
\usepackage{tikz}
\usepackage{textcomp}
\usepackage{lipsum}
\usepackage{multirow}
\usepackage{pifont}
\usepackage{booktabs}
\usepackage{afterpage}
\usetikzlibrary{positioning, arrows.meta, spy, calc}

\usepackage{footnote}
\usepackage{fancyhdr}
\usepackage[absolute]{textpos}
\newcommand{\cmark}{\ding{51}}%
\newcommand{\xmark}{\ding{55}}%

\acrodef{dataset}[GeGrounD]{Geospatial Ground Dataset}
\acrodef{method}[GrounDiff]{Ground Diffusion}
\acrodef{stitching}[PrioStitch]{Prior-Guided Stitching}
\acrodef{ALS}[ALS]{Airborne Laser Scanning}
\acrodef{DOP}[DOP]{\emph{Digital Orthophoto}}
\acrodef{DTM}[DTM]{\emph{Digital Terrain Model}}
\acrodef{DSM}[DSM]{\emph{Digital Surface Model}}
\acrodef{nDSM}[nDSM]{\emph{Normalized Digital Surface Model}}

\acrodef{SfM}[SfM]{\emph{Structure from Motion}}
\acrodef{BEV}[BEV]{\emph{Bird’s Eye View}}
\acrodef{LiDAR}[LiDAR]{\emph{Light Detection and Ranging}}
\acrodef{NURBS}[NURBS]{\emph{Non-Uniform Rational B-Splines}}
\acrodef{LoD}[LoD]{\emph{Levels of Detail}}
\acrodef{SLAM}[SLAM]{\emph{Simultaneous Localization and Mapping}}
\acrodef{MVS}[MVS]{\emph{Multi-view stereo}}
\acrodef{TSGF}[TSGF]{\emph{Two-Steps Semi-Global Filtering}}
\acrodef{Nerf}[Nerf]{\emph{Neural Radiance Fields}}
\acrodef{DEM}[DEM]{\emph{Digital Elevation Model}}
\acrodef{PMF}[PMF]{\emph{Progressive Morphological Filter}}
\acrodef{DRM}[DRM]{\emph{Digital Road Model}} 
\acrodef{PMHR}[PMHR]{\emph{Improved Progressive Morphological Filter based on Hierarchical Radial Basis Function Interpolation}}
\acrodef{TIN}[TIN]{\emph{Triangulated Irregular Network}}
\acrodef{RGT}[RGT]{\emph{Regular Grid Trianglation}}
\acrodef{MLP}[MLP]{\emph{Multi-Layer Perceptron}}
\acrodef{RMSE}[RMSE]{\emph{Root Mean Squared Error}}
\acrodef{MAE}[MAE]{\emph{Mean Absolute Error}}
\acrodef{MAD}[MAD]{\emph{Mean Absolute Deviation}}
\acrodef{MED}[MED]{\emph{Mean Euclidean Distance}}
\acrodef{SMRF}[SMRF]{\emph{Simple Morphological Filter}}
\acrodef{SSIM}[SSIM]{\emph{Structural Similarity Index Measure}}
\acrodef{FILTER}[ECSRC]{\emph{Elevation-Constrained Spatial Road Clustering}}
\acrodef{RANSAC}[RANSAC]{\emph{Random Sample Consensus}}
\acrodef{DSC-SIFI}[SIFI]{Fabulous Sindelfingen}
\acrodef{DSC-STR}[STR]{Stunning Stuttgart}
\acrodef{DSC-MUC}[MUC]{Great Munich}
\acrodef{DSC-BER}[BER]{Vibrant Berlin}
\acrodef{RTK}[RTK]{Real-Time Kinematic Positioning}
\acrodef{GNSS}[GNSS]{Global Navigation Satellite System}
\acrodef{IMU}[IMU]{Inertial Measurement Unit}
\acrodef{GCP}[GCP]{Ground Control Point}
\acrodef{UAV}[UAV]{Unmanned Aerial Vehicle}
\acrodef{DSC}[DSC3D]{DeepScenario Open 3D Dataset}
\acrodef{CSF}[CSF]{Cloth Simulation Filtering}
\acrodef{SBM}[SBM]{Skewness Balancing Method}
\acrodef{PTD}[PTD]{Progressive \ac{TIN} densification}
\acrodef{DALES}[DALES]{Dayton Annotated LiDAR Earth Scan}
\acrodef{NB}[NB]{New Brunswick}

\newcommand{\img}[1]{
  \setlength{\fboxsep}{0pt}
  \setlength{\fboxrule}{0.5pt}
  \fbox{\includegraphics[width=\linewidth]{#1}}%
}

\newcommand{\barelement}[3]{%
  \resizebox{\linewidth}{!}{
    \begin{tabular}{c}
      \setlength{\fboxsep}{0pt}%
      \setlength{\fboxrule}{0.3pt}%
      \fbox{\includegraphics[width=\linewidth,height=0.12cm]{#1}}%
      \\[-5.5pt]
      \makebox[\dimexpr\linewidth+2\fboxrule\relax][s]{%
        \rule{0.3pt}{8pt}\hfill\rule{0.3pt}{8pt}}%
      \\[-5pt]
      \makebox[\dimexpr\linewidth+2\fboxrule\relax][s]{\scriptsize \, #2 \hfill #3\,\,}%
    \end{tabular}%
  }%
}

\newcommand{\ImgWithValue}[3][]{%
\tikz[baseline]{%
    \node[anchor=south west,inner sep=0] (img) {\includegraphics[width=#2]{#3}}; 
    \ifx&#1&\else
        \node[anchor=north east, font=\scriptsize, text=white, fill=black, opacity=0.8, inner sep=1pt] 
        at (img.north east) {\textbf{#1}};
    \fi
}%
}

\def\wacvPaperID{2918} 
\def\confName{WACV}
\def\confYear{2026}

\title{GrounDiff: Diffusion-Based Ground Surface Generation\\from Digital Surface Models}

\makeatletter
\renewcommand{\@fnsymbol}[1]{}
\makeatother

\author{Oussema Dhaouadi$^{1,2,3,\dagger}$ \quad Johannes Meier$^{1,2,3}$ \quad Jacques Kaiser$^{1}$ \quad Daniel Cremers$^{2,3}$ \\
$^{1}$ DeepScenario\quad $^{2}$ TU Munich\quad $^{3}$ Munich Center for Machine Learning
\thanks{\scriptsize $\dagger$ Corresponding author.}%
\thanks{\scriptsize TUM: {\tt \{oussema.dhaouadi, j.meier, cremers\}@tum.de}}
\thanks{\scriptsize DeepScenario: \tt jacques@deepscenario.com}
}

\begin{document}
\maketitle
\begin{abstract}
Digital Terrain Models (DTMs) represent the bare-earth elevation and are important in numerous geospatial applications. Such data models cannot be directly measured by sensors and are typically generated from Digital Surface Models (DSMs) derived from LiDAR or photogrammetry. Traditional filtering approaches rely on manually tuned parameters, while learning-based methods require well-designed architectures, often combined with post-processing. To address these challenges, we introduce Ground Diffusion (GrounDiff), the first diffusion-based framework that iteratively removes non-ground structures by formulating the problem as a denoising task. We incorporate a gated design with confidence-guided generation that enables selective filtering. To increase scalability, we further propose Prior-Guided Stitching (PrioStitch), which employs a downsampled global prior automatically generated using GrounDiff to guide local high-resolution predictions. We evaluate our method on the DSM-to-DTM translation task across diverse datasets, showing that GrounDiff consistently outperforms deep learning-based state-of-the-art methods, reducing RMSE by up to 93\% on ALS2DTM and up to 47\% on USGS benchmarks. In the task of road reconstruction, which requires both high precision and smoothness, our method achieves up to 81\% lower distance error compared to specialized techniques on the GeRoD benchmark, while maintaining competitive surface smoothness using only DSM inputs, without task-specific optimization. Our variant for road reconstruction, GrounDiff+, is specifically designed to produce even smoother surfaces, further surpassing state-of-the-art methods. The project page is available at \href{https://deepscenario.github.io/GrounDiff/}{deepscenario.github.io/GrounDiff}.
\end{abstract}    
\section{Introduction}
\label{sec:introduction}
\begin{figure}[t]
  \centering
  \includegraphics[width=\linewidth]{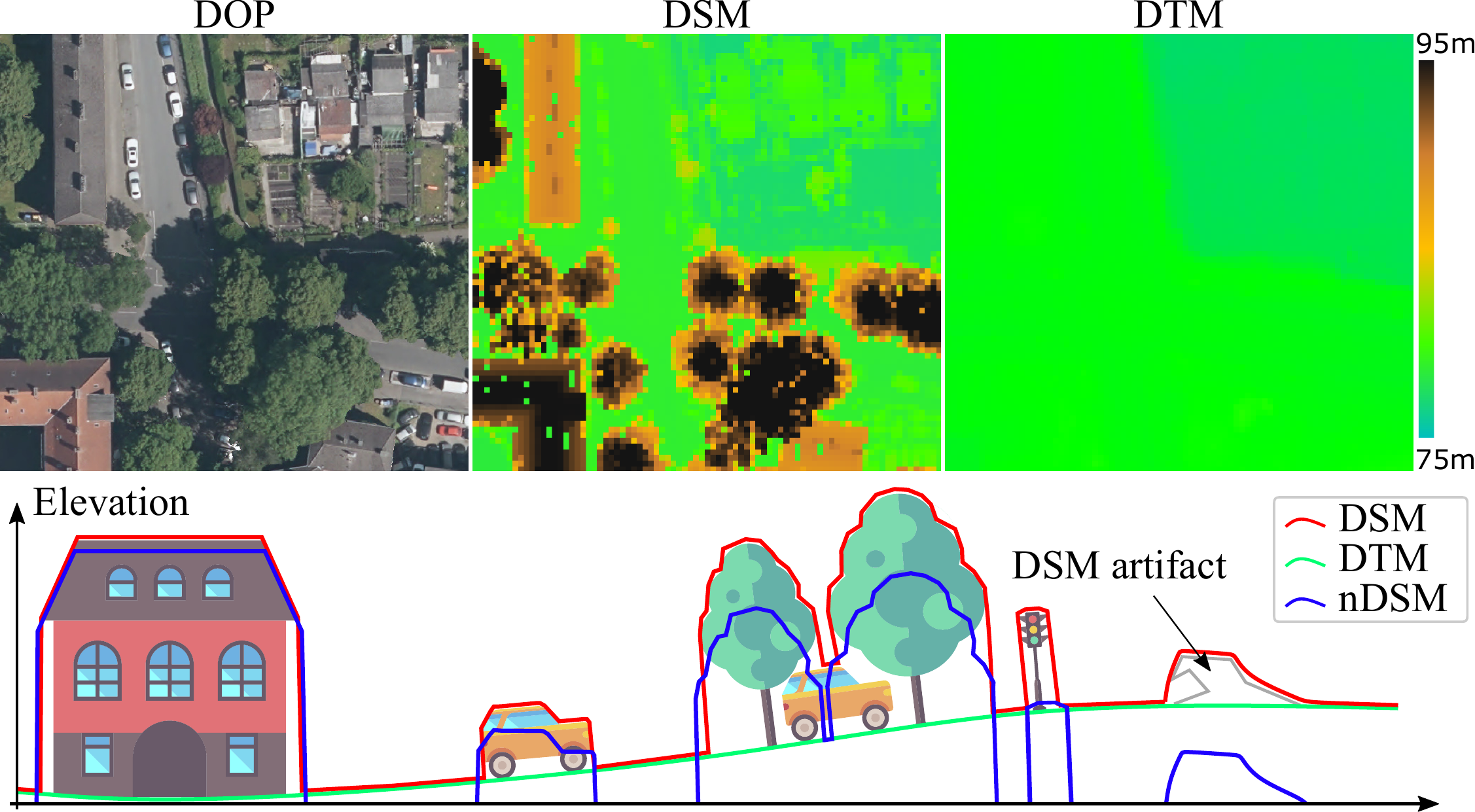}
  \caption{\textbf{Geospatial surface models.} Comparison between \ac{DSM}, \ac{DTM}, and \ac{nDSM}.}
  \label{fig:dsm_dtm}
  \vspace*{-1.25\baselineskip}
\end{figure}

\begin{figure}[t]
    \centering
    \includegraphics[width=\linewidth]{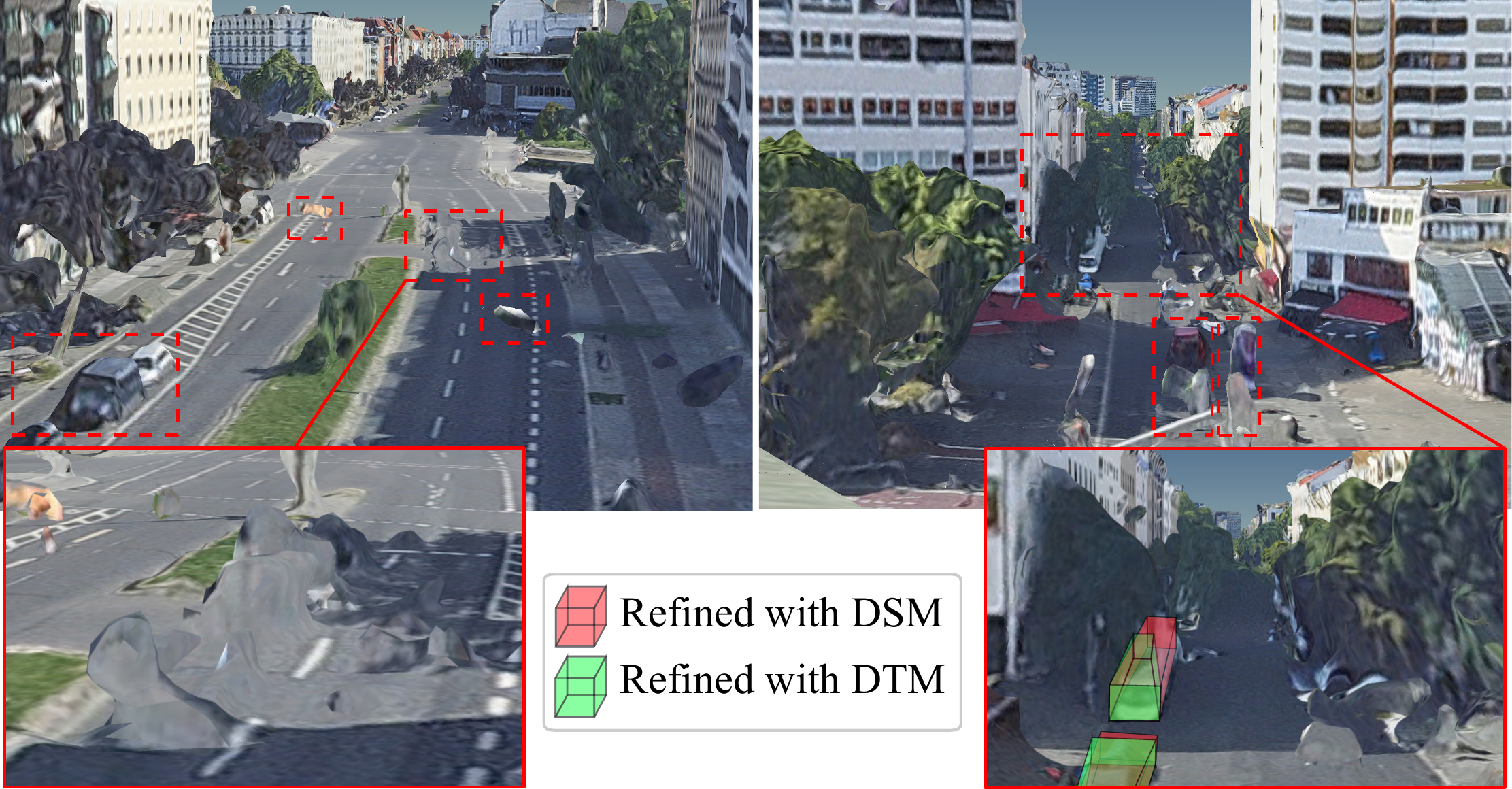}
    \caption{%
    \textbf{\ac{DTM} applications in autonomous driving: object detection refinement using geospatial data.} 
Left: Textured 3D mesh with surface noise and artifacts affecting \ac{DSM} quality. 
Right: 3D bounding box height refinement via raycasting---the red box (using noisy \ac{DSM}) shows incorrect vertical positioning due to surface artifacts, while the green box (using clean \ac{DTM}) achieves accurate ground-level placement essential for safe navigation.\protect\footnotemark}
    \label{fig:rec_quality}
    \vspace{-1\baselineskip}
\end{figure}

\acp{DSM} are 2.5D raster representations capturing surface elevations including vegetation and man-made structures, derived from airborne LiDAR~\cite{dong2017lidar} or photogrammetry~\cite{forstner2016photogrammetric}. \acp{DTM} represent the underlying bare-earth surface with above-ground objects removed, while the non-ground relative elevation data is represented by \ac{nDSM}, as illustrated in \cref{fig:dsm_dtm}. This distinction is crucial for numerous applications: infrastructure planning~\cite{xu2020road}, autonomous navigation~\cite{toscano2023aia,domiciano2024use}, flood modeling~\cite{muthusamy2021understanding}, forest management~\cite{niemi2023using}, precision agriculture~\cite{erunova2024geomorphometric}, and geological analysis~\cite{fajri2019lineament}. \cref{fig:rec_quality} shows a usage example in 3D detection.
Extracting \acp{DTM} from \acp{DSM} is challenging, especially in steep terrain, dense vegetation, or large urban areas. Traditional filtering approaches~\cite{zhang2003progressive,axelsson2000generation_lastools,zhang2016easy} rely on manually tuned parameters that often fail in heterogeneous landscapes and struggle with scalability across different terrain types. Recent deep learning methods~\cite{amirkolaee2022dtm,bittner2023dsm2dtm,naeini2024ressub} show promise but suffer from limited generalization to complex scenarios, and often require extensive post-processing.

Diffusion models have revolutionized generative modeling through iterative denoising~\cite{ho2020denoising,dhariwal2021diffusion}. This paradigm naturally aligns with \ac{DTM} extraction, where above-ground structures can be conceptualized as noise to be systematically removed while preserving terrain. We introduce \ac{method}, the first diffusion-based \ac{DTM} extraction approach that progressively removes non-ground structures while maintaining topographic details.

To address large-scale terrain modeling limitations of current approaches, we develop \ac{stitching}, a scalable processing strategy that first generates a low-resolution \ac{DTM} estimate serving as a conditional prior for high-resolution tile processing. Moving window blending techniques then stitch these tiles, enabling processing of arbitrarily large areas while maintaining local detail, which is essential for real-world deployment across extensive geographic regions.

To summarize, our main contributions include:

\begin{itemize}
    \item \ac{method}, a novel diffusion framework for \ac{DSM}-to-\ac{DTM} conversion progressively denoising above-ground structures.
    
    \item \ac{stitching}, a scalable processing strategy through low-resolution \ac{DTM} conditioning and tile blending.
    
    \item Comprehensive evaluation across USGS~\cite{su_2020,rt_2018,kw_2020}, ALS2DTM~\cite{le2022learning}, and GeRoD~\cite{flexroad} datasets along with ablation studies, outperforming state-of-the-art methods.
\end{itemize}

\footnotetext{Data source: 3D City Model of Berlin, © Berlin Partner für Wirtschaft und Technologie GmbH.}

\section{Related Work}
\label{sec:related_work}
\subsection{DTM Generation}
\label{subsec:dtm_generation}
\ac{DTM} generation involves recovering bare-earth surfaces from elevation data containing above-ground structures. Approaches range from traditional filtering methods to learning-based techniques, each addressing specific challenges.
\subsubsection{Traditional Methods}
Classical \ac{DTM} extraction follows a two-stage pipeline: ground filtering to identify terrain points, followed by surface interpolation~\cite{hutchinson2011recent}. These methods broadly fall into three categories.

\textbf{Morphological filtering} applies structuring elements to identify elevation outliers based on local terrain characteristics. The \ac{PMF}~\cite{zhang2003progressive} and \ac{SMRF}~\cite{pingel2013improved} are representative approaches. Extensions address scale sensitivity through multi-scale operations~\cite{duan2019large} or directional filtering~\cite{pijl2020terra}.

\textbf{Statistical approaches} aim to reduce parameter sensitivity by leveraging data-driven criteria. \ac{SBM}~\cite{bartels2010threshold} applies statistical thresholds and relies on strong assumptions of relatively flat ground, whereas \ac{CSF}~\cite{zhang2016easy} relies on physics-based modeling and has gained widespread adoption due to its robustness. Despite its popularity, \ac{CSF} suffers from limitations such as loss of ground adhesion on rising terrain, spurious sinks caused by correlation artifacts in \ac{DSM} computation, and poor scalability for large-scale data.

\textbf{Surface-based methods} reconstruct terrain through geometric surface modeling. \ac{PTD}~\cite{axelsson2000generation_lastools} iteratively grows sparse seed triangulations, with recent variants addressing parameter tuning~\cite{shi2018parameter} and adaptive thresholding~\cite{cai2019filtering,zheng2024improved}. FlexRoad~\cite{flexroad} reconstructs road surfaces by fitting parametric \ac{NURBS} to \ac{DSM} regions segmented using \ac{DOP} data. Though effective for smoothness-critical applications, it requires hyperparameter tuning and auxiliary data, limiting large-scale deployment. Its \ac{NURBS}-based approach struggles with complex elevation transitions such as tunnels and bridges, while online surface-fitting constrains computational efficiency for on-device solutions.

Despite their maturity, traditional methods require terrain-specific parameter tuning and often fail in complex environments with dense vegetation, dense urban regions, or steep topography. Our approach specifically addresses these limitations.

\begin{figure*}[!tb]
\centering
\includegraphics[width=1\linewidth]{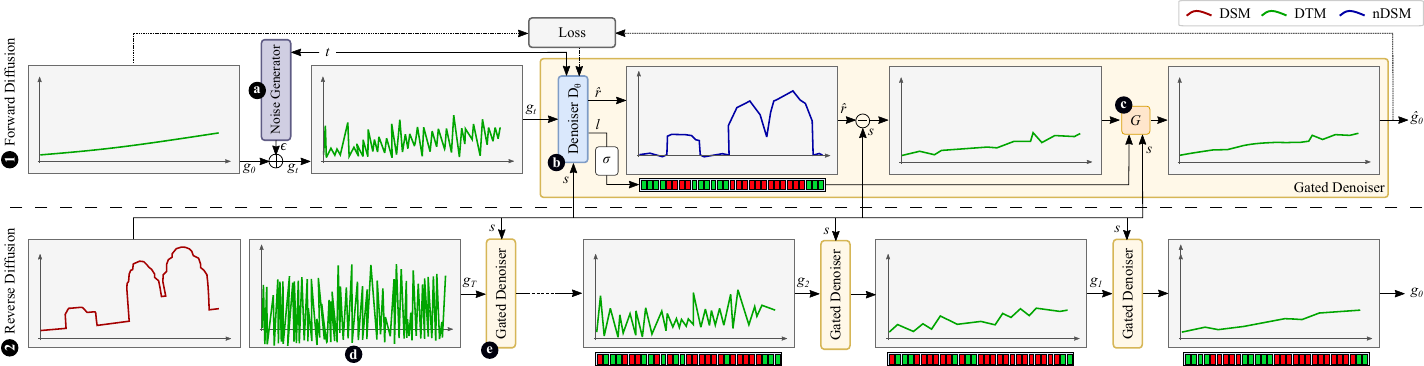}
\caption{\textbf{Method overview (on 1D terrain for clarity).} (1) Training: Forward diffusion process where (a) ground-truth \ac{DTM} $g_0$ is corrupted with noise to obtain $g_t$, and (b) denoiser takes noisy terrain $g_t$ and \ac{DSM} $s$ to predict \ac{nDSM} $\hat{r}$ and classification logits $l$. The \ac{nDSM} is subtracted from \ac{DSM} to generate initial \ac{DTM}, then (c) refined using ground probabilities to produce final estimate $\hat{g}_0$. (2) Inference: Reverse process starts with (d) prior (e.g., Gaussian noise, noisy \ac{DSM}, or low-resolution \ac{DTM}) and iteratively applies the gated denoiser (e) conditioned on \ac{DSM} $s$, progressively denoising from $g_T$ to $g_0$ to recover the final \ac{DTM}.}
\label{fig:overview}
\vspace*{-1.25\baselineskip}
\end{figure*}
\subsubsection{Learning-Based Methods}
Deep learning approaches aim to capture terrain priors directly from data, reducing manual parameter tuning.

\textbf{Classification formulations} treat \ac{DTM} generation as pixel-wise ground/non-ground segmentation~\cite{hu2016deep,gevaert2018deep,yang2018segmentation}. These methods apply CNNs to height images, then interpolate classified ground points to generate continuous surfaces. Štroner \etal \cite{vstroner2025multi} directly uses fully connected layers in a triangular shape. However, interpolation introduces artifacts and loses fine terrain structure~\cite{amirkolaee2022dtm}, resulting in oversmoothed terrain or terrain with artifacts.

\textbf{Regression formulations} directly predict terrain heights, avoiding interpolation bottlenecks. Recent approaches include GANs for \ac{DSM}-to-\ac{DTM} translation~\cite{le2022learning,oshio2023generating}, multi-scale fusion networks~\cite{amirkolaee2022dtm}, and U-Net architectures with EfficientNet encoders~\cite{bittner2023dsm2dtm}. RESSUB-Net~\cite{naeini2024ressub} explicitly models elevation residuals, while physics-informed autoencoders incorporate geometric priors~\cite{alizadeh2024advancing}.

These methods move toward end-to-end learning of flexible terrain representations that generalize across environments, yet they struggle with spatial consistency in complex terrain and large-scale areas. Our diffusion-based \ac{method} combines confidence-guided regression with ground mask prediction, while our tiling strategy \ac{stitching} addresses scalability.

\subsection{Diffusion Models for Image Translation}
Diffusion models have emerged as state-of-the-art generative frameworks, excelling in image synthesis and translation tasks~\cite{ho2020denoising,dhariwal2021diffusion}. By learning to iteratively reverse noise injection processes, they capture complex distributions with superior stability compared to GANs.
Conditional variants enable guided generation using auxiliary inputs and have been adapted to numerous tasks. SR3~\cite{saharia2022image} achieves high-quality super-resolution through iterative denoising, while Stable Diffusion~\cite{rombach2022high} combines U-Net backbones with cross-attention for scalable conditional synthesis. Palette~\cite{saharia2022palette} demonstrates versatile image-to-image translation across colorization, inpainting, and restoration tasks using a single unified architecture. 
Despite their proven effectiveness in image translation, diffusion models remain unexplored for geospatial applications, particularly \ac{DSM}-to-\ac{DTM} conversion. By treating above-ground structures as noise to be iteratively removed, diffusion models offer a natural framework for terrain extraction that aligns with the denoising paradigm. This observation motivates our proposed \ac{method} approach.

\section{Method}
\label{sec:method}

We address the \ac{DSM}-to-\ac{DTM} challenge through (1)~\ac{method}, a diffusion-based framework that reformulates terrain extraction as denoising, and (2)~\ac{stitching}, a strategy that uses \ac{method} to process large-scale \acp{DSM}.

\subsection{Problem Formulation}
We formulate \ac{DSM}-to-\ac{DTM} conversion as a conditional diffusion process. The \ac{nDSM} residual $r = s - g$ captures non-ground structures, where $s$ denotes the \ac{DSM} and $g$ the target \ac{DTM}. Our approach employs a forward corruption that progressively perturbs the terrain and a conditional reverse generation that reconstructs its clean version. \cref{fig:overview} illustrates this approach. For clarity, we visualize 1D elevation profiles representing a cross-sectional view of the 2D elevation maps.

\begin{figure*}[!tb]
\centering
\includegraphics[width=1\linewidth]{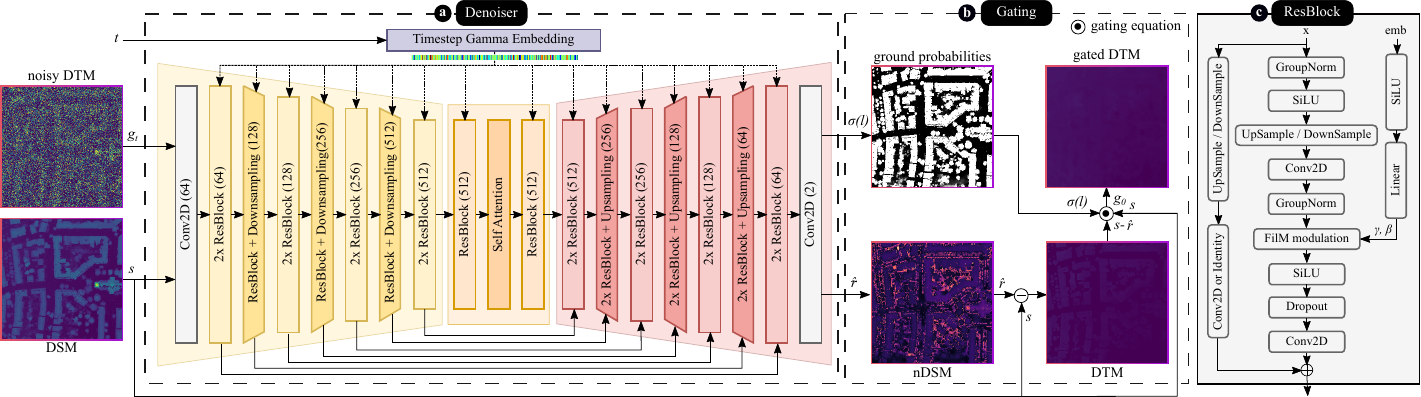}
\caption{\textbf{Our denoiser architecture.} (a)~The network follows a U-Net encoder–decoder based on (c)~residual blocks with FiLM~\cite{perez2018film} timestep conditioning, and skip connections. Dual outputs produce residual corrections $\hat{r}$ and confidence logits $\ell$, which are (b)~combined via the gating function in \cref{eq:gating}.}
\label{fig:unet}
\vspace*{-1.25\baselineskip}
\end{figure*}

\subsection{Diffusion-Based Ground Surface Generation}

\paragraph{Forward diffusion process.}
The forward process corrupts clean terrain $g_0 = g$ through latent states $\{g_t\}_{t=1}^T$ using variance schedule $\{\beta_t\}$, where $\alpha_t = 1 - \beta_t$ and $\bar{\alpha}_t = \prod_{i=1}^t \alpha_i$. The marginal distribution is:
\begin{align}
q(g_t \mid g_0) &= \mathcal{N}\!\left(g_t \;\middle|\; \sqrt{\bar{\alpha}_t}\, g_0,\; (1 - \bar{\alpha}_t)\mathbf{I}\right),
\end{align}
equivalently expressed as:
\begin{align}
g_t &= \sqrt{\bar{\alpha}_t}\, g_0 + \sqrt{1 - \bar{\alpha}_t}\, \epsilon, \quad \epsilon \sim \mathcal{N}(0, \mathbf{I}).
\end{align}
At each step, we compute a gated \ac{DTM} estimate:
\begin{align}
\hat{g}_0 &= G(D_\theta(g_t, s, t), s),
\end{align}
where $D_\theta$ is our denoiser network and $G$ the gating function detailed in the following section.

\paragraph{Denoiser network.}
Our denoiser $D_\theta$ has a U-Net backbone inspired by~\cite{ho2020denoising}, with multiple adaptations. The architecture uses residual and attention blocks and integrates timestep conditioning, as shown in \cref{fig:unet}. Given corrupted \ac{DTM} $g_t$, \ac{DSM} $s$, and diffusion timestep $t$, the model outputs residual correction $\hat{r}$ and per-pixel confidence logits~$\ell$:
\begin{align}
(\hat{r}, \ell) &= D_\theta(g_t, s, t).
\end{align}

The inputs are concatenated channel-wise as $[g_t, s]$ and passed through a convolutional stem before encoder processing. The encoder progressively downsamples features using ResBlocks. A bottleneck stage aggregates global context via a residual–attention–residual stack, where spatial self-attention captures dependencies across terrain structures. The decoder mirrors the encoder with upsampling, internal residual skip connections, and external skip connections from the encoder, effectively fusing high-level semantic context with fine-grained spatial detail. Timestep embeddings are injected into each ResBlock via FiLM~\cite{perez2018film} modulation, enabling the network to adapt its internal representation to the current diffusion step.

The output head branches into two maps: (i) residual prediction $\hat{r}$ representing correction signal to the noisy input terrain, and (ii) per-pixel logits $\ell$ that estimate confidence for ground classification. This dual-output design leverages the observation that \ac{DSM} and \ac{DTM} are highly correlated in ground-visible regions, where minimal correction is needed. To ensure modifications are applied only where necessary, we introduce a gating mechanism that selectively fuses the \ac{DSM} $s$ with the residual correction:
\begin{align}
G(\hat{r}, \ell, s) &= \sigma(\ell) \odot s + \big(1 - \sigma(\ell)\big) \odot (s - \hat{r}),
\label{eq:gating}
\end{align}
where $\odot$ denotes element-wise multiplication and $\sigma$ is the sigmoid function. This segmentation-based fusion enables the network to preserve \ac{DSM} values in confidently classified ground regions while focusing computational effort on interpolating terrain in regions where ground is occluded by non-terrain structures.

\paragraph{Reverse diffusion process.}
The reverse process reconstructs $g_0$ through iterative denoising over $T$ steps. At each step $t$, we sample from:
\begin{align}
p(g_{t-1} \mid g_t, s) &= \mathcal{N}\!\left(g_{t-1} \;\middle|\; \mu_\theta(g_t, s, t), \sigma_t^2 \mathbf{I}\right),
\end{align}
with mean and variance:
\begin{align}
\hat{g}_t &= G(D_\theta(g_t, s, t), s),\\
\mu_\theta(g_t, s, t) &= \frac{\beta_t \sqrt{\bar{\alpha}_{t-1}}}{1 - \bar{\alpha}_t}\,\hat{g}_t + \frac{(1 - \bar{\alpha}_{t-1}) \sqrt{\alpha_t}}{1 - \bar{\alpha}_t}\, g_t, \\
\sigma_t^2 &= \frac{\beta_t (1 - \bar{\alpha}_{t-1})}{1 - \bar{\alpha}_t}.
\end{align}
Sampling from the posterior distribution yields:
\begin{align}
g_{t-1} &= \mu_\theta(g_t, s, t) + \sigma_t \, \epsilon, \quad \epsilon \sim \mathcal{N}(0, \mathbf{I}).
\end{align}
Starting from Gaussian noise $g_T \sim \mathcal{N}(0, \mathbf{I})$, this process iteratively generates the final \ac{DTM} estimate $g_0$. We show that initializing $g_T \sim \mathcal{N}(s, \mathbf{I})$, which combines structural information with stochastic noise, improves the efficiency of the generation process.

\paragraph{Training objective.}
\label{sec:training_losses}
Given ground-truth \ac{DTM} $g$, predicted \ac{DTM} $\hat{g}$, and logits $\ell$, we optimize a multi-component loss function combining regression, edge-aware, and classification terms:
\begin{align}
\label{eq:loss}
\mathcal{L} &= \lambda_1\mathcal{L}_{1} + \lambda_2\mathcal{L}_{2} + \lambda_{\nabla}\mathcal{L}_{\nabla} + \lambda_{c}\mathcal{L}_{c}.
\end{align}
The regression terms are:
\begin{align}
\mathcal{L}_{1} &= \|\hat{g} - g\|_1, 
\quad 
\mathcal{L}_{2} = \|\hat{g} - g\|_2^2,
\end{align}
where $\mathcal{L}_2$ smooths homogeneous regions such as roads or areas beneath buildings, while $\mathcal{L}_1$ preserves sharp ground transitions along building edges or abrupt terrain features. Earlier works relied solely on $\mathcal{L}_1$ regression losses \cite{le2022learning,bittner2023dsm2dtm}, whereas \cite{amirkolaee2022dtm,naeini2024ressub} extended this with additional geometric terms based on gradient components and normalized normals. In contrast, we adopt a simplified edge-aware regularization that focuses only on gradient magnitudes:
\begin{align}
\mathcal{L}_{\nabla} &= \big\| \, \lVert \nabla \hat{g} \rVert_2 - \lVert \nabla g \rVert_2 \, \big\|_1.
\end{align}

This magnitude-only formulation is advantageous since ground-truth \acp{DTM} often use triangulation beneath non-ground structures, creating arbitrary orientation patterns. By focusing on gradient magnitudes rather than full vectors, our model maintains terrain roughness while avoiding overfitting to these artificial interpolations, producing more realistic reconstructions. Finally, the confidence head employs binary cross-entropy loss:
\begin{align}
\mathcal{L}_{c} &= \text{BCE}(\sigma(\ell), M_\alpha),
\end{align}
where $M_{\alpha}$ indicates ground pixels satisfying $|r| \nobreak<\nobreak \alpha$.

\subsection{Scaling to Large-Scale DTMs}
While \ac{method} is trained on fixed-size \ac{DSM} patches matching the network's input dimensions, real-world \acp{DSM} often span kilometers at high resolution. Direct application on large inputs leads to memory constraints and potential inconsistencies. To address this, we introduce \ac{stitching}, a prior-guided tiling strategy that enables coherent \ac{DTM} generation for arbitrarily large areas.

\begin{figure}[t]
    \centering
    \includegraphics[width=\linewidth]{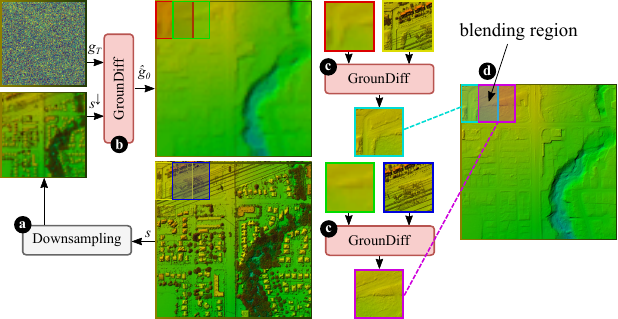}
    \caption{\textbf{\ac{stitching} strategy.} Our approach scales \ac{method} to large \acp{DSM} through: low-resolution prior generation by (a) downsampling the input \ac{DSM} and (b) applying \ac{method}, (c) tiling the original \ac{DSM} into overlapping patches, conditioning each patch with the corresponding region from the upsampled prior \ac{DTM}, and (d) blending the processed tiles using weighted fusion to produce the final high-resolution \ac{DTM}.}
    \label{fig:priostitch}
    \vspace*{-1.25\baselineskip}
\end{figure}

\ac{stitching} operates in a coarse-to-fine manner, as illustrated in \cref{fig:priostitch}. We begin with \textbf{generating a global prior}, where we downsample the input \ac{DSM} to the network's input dimensions and process it through \ac{method} to produce a global prior \ac{DTM} that captures low-frequency terrain structure. This prior provides a coherent baseline for subsequent refinement steps.

Next, we perform \textbf{tile extraction} by dividing the original high-resolution \ac{DSM} into overlapping tiles that match the network's input size. The overlap margin is configurable, allowing flexibility in processing.

Then, we apply \textbf{conditional refinement} for each tile by extracting the corresponding region from the upsampled prior \ac{DTM}. Instead of initializing the diffusion process with Gaussian noise, we directly provide this prior \ac{DTM} as the initial state for the denoiser. This follows the training formulation, which uses noisy ground-truth \acp{DTM} as input. Conditioning in this way significantly improves consistency and accuracy, guiding the generation process with global context and allowing \ac{method} to maintain coherence across large areas while focusing on detail refinement.

Finally, we perform \textbf{weighted blending} to merge the overlapping tile outputs. We use strategies such as averaging, minimum or maximum value selection, and weighted functions including linear, cosine, or exponential decay.
\section{Experiments}
We evaluate \ac{method} on ground surface generation and road reconstruction across diverse environments and multiple benchmark datasets.

\subsection{Experimental Setup}

\paragraph{Datasets.}
We evaluate our method on three benchmark collections: (1)~\textit{USGS (OpenTopography)}, which includes five American datasets (SU-I, SU-II, SU-II~\cite{su_2020}, RT~\cite{rt_2018}, and KW~\cite{kw_2020}) encompassing diverse landscapes such as urban, residential, mountainous, and coastal regions; (2)~\textit{ALS2DTM}~\cite{le2022learning}, comprising two Canadian datasets representing residential and forested landscapes: DALES (urban) and NB (suburban and rural); and (3)~\textit{GeRoD}~\cite{flexroad} providing German urban and highway data designed for road reconstruction benchmarking. Further details on the datasets are provided in the supplementary material.

\paragraph{Baselines.}
For \textit{ground generation}, we evaluate \ac{method} against established approaches including traditional filtering methods: \ac{PMF}~\cite{zhang2003progressive} (progressive morphological filtering), \ac{SBM}~\cite{bartels2010threshold} (slope-based morphology with adaptive thresholds), \ac{SMRF}~\cite{pingel2013improved} (morphological filtering with progressive windows), \ac{CSF}~\cite{zhang2016easy} (physics-based cloth simulation), and \ac{PTD}~\cite{axelsson2000generation_lastools} (progressive TIN densification). Learning-based approaches include DeepTerRa~\cite{le2022learning} (GAN-based DSM-to-DTM translation), HDCNN~\cite{amirkolaee2022dtm} (hierarchical CNN with multi-scale fusion), and RESSUB-Net~\cite{naeini2024ressub} (residual U-Net modeling elevation differences).

For \textit{road reconstruction}, we compare our method against plane fitting (a simple planar approximation), \ac{RGT} (which interpolates gaps in the \acp{DSM} segmented for roads using \ac{DOP} through triangulation), and FlexRoad~\cite{flexroad} (which performs parametric \ac{NURBS} fitting using the same segmented \acp{DSM} as \ac{RGT}).

\paragraph{Evaluation metrics.}
For \textit{ground generation}, we use regression metrics including \ac{RMSE} and \ac{MAE} in meters to measure the elevation accuracy. Classification performance is assessed by Type~I error $E_{T_1}$ (retaining non-ground points), Type~II error $E_{T_2}$ (removing ground points), and total error $E_{tot}$ (sum of both), all expressed as percentages.

For \textit{road reconstruction}, following~\cite{flexroad}, accuracy is measured using \ac{MED} between reconstructed surfaces and ground-truth road and non-road point clouds, while surface smoothness quality is quantified by \ac{MAD}.

\subsection{Ground Generation Results}

Our evaluation follows a twofold approach: first, comparing against DeepTerRa~\cite{le2022learning} on ALS2DTM datasets, and second, comparing against HDCNN~\cite{amirkolaee2022dtm} and RESSUB-Net~\cite{naeini2024ressub} on USGS datasets.

\begin{table}[ht]
\centering
\footnotesize
\begin{tabular}{lcc}
\toprule
\bfseries{Method} & \bfseries{DALES}$\downarrow$ & \bfseries{NB}$\downarrow$ \\
\midrule
\textit{Traditional Methods} & & \\
\quad \ac{PMF}~\cite{zhang2003progressive} & 4.27 & 0.81 \\
\quad \ac{SBM}~\cite{bartels2010threshold} & 3.59 & 4.15 \\
\quad \ac{SMRF}~\cite{pingel2013improved} & 4.08 & \underline{0.75} \\
\quad \ac{CSF}~\cite{zhang2016easy} & 1.19 & 0.88 \\
\midrule
\textit{Learning-based Methods} & & \\
\quad DeepTerRa${}^{\ast}$~\cite{le2022learning} & 6.31 & 6.76 \\
\quad DeepTerRa${}^{\dagger}$~\cite{le2022learning} & \underline{0.82} & 0.98 \\
\quad \ac{method}~{\scriptsize (Ours)} & \textbf{0.51} & \textbf{0.45} \\
\bottomrule
\end{tabular}

\caption{\textbf{Ground generation results on ALS2DTM~\cite{le2022learning} datasets.} Performance comparison using RMSE in meters (lower is better) across two distinct test regions: DALES (urban) and NB (suburban and rural). \ac{method} achieves superior performance against both traditional and learning-based methods. DeepTerRa${}^*$ uses only \ac{DSM} input. DeepTerRa${}^{\dagger}$ uses lower/mean/upper elevation rasters, height statistics, and semantic information as inputs. \textbf{Bold}: best performance, \underline{underlined}: second best.}
\label{tab:regression_als2dtm}
\vspace*{-1.25\baselineskip}
\end{table}

\paragraph{ALS2DTM benchmarks.}
\cref{tab:regression_als2dtm} reports results on DALES and NB. \ac{method} achieves the lowest RMSE on both datasets. On DALES, it reaches 0.51\,m, a 38\% reduction compared to DeepTerRa$^{\dagger}$ at 0.82\,m, which was the previous best method and relies on multiple inputs including elevation statistics and semantic channels. On NB, our method attains 0.45\,m, improving by 54\% over DeepTerRa$^{\dagger}$ at 0.98\,m and by 40\% over SMRF at 0.75\,m, the strongest traditional baseline. Under identical input conditions, the contrast is even stronger: DeepTerRa$^{\ast}$, which also uses only DSM, records 6.31\,m on DALES and 6.76\,m on NB, while our method reduces these errors to 0.51\,m and 0.45\,m, corresponding to reductions of 92\% and 93\% respectively. These results show that the improvements stem from the modeling approach itself rather than from additional modalities, enabling state-of-the-art \ac{DTM} generation from \ac{DSM} input alone.

\begin{table}[ht]
\centering
\setlength{\tabcolsep}{6pt}
\footnotesize
\begin{tabular}{lcccc}
\toprule
\bfseries{Method} & \bfseries{RMSE}$\downarrow$ & \bfseries{$\boldsymbol{E_{T_1}}$}$\downarrow$ & \bfseries{$\boldsymbol{E_{T_2}}$}$\downarrow$ & \bfseries{$\boldsymbol{E_{tot}}$}$\downarrow$ \\
\midrule
\textit{SU-II Dataset}~\cite{su_2020} & & & & \\
\quad gLidar~\cite{mongus2014ground_glidar} & 2.110 & 12.56 & 14.55 & 13.22 \\
\quad \ac{PTD} (LAStools)~\cite{axelsson2000generation_lastools} & 0.564 & 9.63 & 14.08 & 10.61 \\
\quad \ac{CSF}~\cite{zhang2016easy} & 4.501 & 19.62 & 18.30 & 18.67 \\
\quad HDCNN~\cite{amirkolaee2022dtm} & 0.281 & \underline{3.34} & \textbf{0.88} & \textbf{1.35} \\
\quad RESSUB-Net~\cite{naeini2024ressub} & \underline{0.178} & \textbf{1.00} & 7.10 & \underline{2.70} \\
\quad \ac{method}~{\scriptsize (Ours)} & \textbf{0.095} & 4.49 & \underline{2.46} & 3.82 \\
\midrule
\textit{SU-III Dataset}~\cite{su_2020} & & & & \\
\quad gLidar~\cite{mongus2014ground_glidar} & 1.325 & 11.18 & 8.91 & 9.84 \\
\quad \ac{PTD} (LAStools)~\cite{axelsson2000generation_lastools} & 2.650 & 21.59 & 9.33 & 14.80 \\
\quad \ac{CSF}~\cite{zhang2016easy} & 1.121 & \textbf{1.20} & 12.11 & 7.72 \\
\quad HDCNN~\cite{amirkolaee2022dtm} & \underline{0.165} & 6.68 & \textbf{0.64} & \textbf{1.29} \\
\quad \ac{method}~{\scriptsize (Ours)} & \textbf{0.099} & \underline{4.72} & \underline{4.53} & \underline{4.17} \\
\midrule
\textit{RT Dataset}~\cite{rt_2018} & & & & \\
\quad gLidar~\cite{mongus2014ground_glidar} & 0.566 & 12.36 & 6.15 & 8.61 \\
\quad \ac{PTD} (LAStools)~\cite{axelsson2000generation_lastools} & 0.742 & 10.73 & 9.45 & 9.21 \\
\quad \ac{CSF}~\cite{zhang2016easy} & 0.721 & \textbf{0.36} & 25.29 & 17.19 \\
\quad HDCNN~\cite{amirkolaee2022dtm} & \underline{0.248} & \underline{0.89} & \textbf{5.24} & \underline{3.61} \\
\quad RESSUB-Net~\cite{naeini2024ressub} & 0.300 & 1.40 & \underline{5.80} & \textbf{2.30} \\
\quad \ac{method}~{\scriptsize (Ours)} & \textbf{0.189} & 8.48 & 7.25 & 4.10 \\
\midrule
\textit{KW Dataset}~\cite{kw_2020} & & & & \\
\quad gLidar~\cite{mongus2014ground_glidar} & 5.871 & 23.91 & 15.21 & 25.61 \\
\quad \ac{PTD} (LAStools)~\cite{axelsson2000generation_lastools} & 16.615 & 32.34 & \textbf{5.13} & 24.83 \\
\quad \ac{CSF}~\cite{zhang2016easy} & 12.000 & \textbf{22.81} & 29.51 & 25.54 \\
\quad HDCNN~\cite{amirkolaee2022dtm} & \textbf{1.636} & \underline{26.80} & \underline{7.12} & 18.11 \\
\quad RESSUB-Net~\cite{naeini2024ressub} & \underline{2.580} & 53.20 & 7.30 & \textbf{15.40} \\
\quad \ac{method}~{\scriptsize (Ours)} & 8.101 & 47.48 & 19.26 & \underline{16.91} \\
\quad ${\text{\ac{method}}}^\dagger$ ~{\scriptsize (Ours)} & 3.301 & 25.81 & 28.58 & 25.15 \\
\bottomrule
\end{tabular}

\caption{\textbf{Ground generation results on USGS datasets.} The metrics include RMSE in meters while classification errors are expressed as percentages. \ac{method} demonstrates superior RMSE across diverse terrain types while maintaining competitive classification performance. \ac{method}${}^\dagger$ is a variant trained on the NB dataset~\cite{le2022learning} instead of SU-I~\cite{su_2020}, in order to match the topology of the test KW dataset~\cite{kw_2020}.
\textbf{Bold}: best performance, \underline{underlined}: second best.}
\label{tab:regression_usgs}
\vspace*{-1.25\baselineskip}
\end{table}

\paragraph{USGS benchmarks.}
Results on the USGS datasets in \cref{tab:regression_usgs} demonstrate \ac{method}'s effectiveness across diverse terrains when only trained on the full SU-I~\cite{su_2020} dataset. Our single-scale approach achieves the lowest \ac{RMSE} on three datasets, outperforming the multi-scale HDCNN by 40\% on SU-III and 24\% on RT, and RESSUB-Net by 47\% on SU-II. Classification metrics remain competitive, with dataset-specific variations in Type~I/II errors. The RT dataset shows higher Type~I error, suggesting more aggressive filtering in this mixed-topology environment, while maintaining lower total error than most traditional methods. The KW dataset remains challenging due to its mountainous terrain with abrupt elevation changes and extreme differences from the training data, highlighting potential for future domain adaptation techniques. The last row shows that another training from scratch on topologically similar data, such as the NB dataset~\cite{le2022learning}, reduces the RMSE by 60\%.

\begin{figure}[ht]
\centering
\setlength{\tabcolsep}{1pt}     
\renewcommand{\arraystretch}{0.5}
\resizebox{\linewidth}{!}{%
\begin{tabular}{@{} >{\centering\arraybackslash}m{0.55cm} *{3}{>{\centering\arraybackslash}m{0.29\linewidth}} @{}}

& {\small SU-II~\cite{su_2020}} 
& {\small RT~\cite{rt_2018}} 
& {\small KW~\cite{kw_2020}} \\
\cmidrule(lr){2-4}

\rotatebox{90}{\small Ground Prob.} &
\img{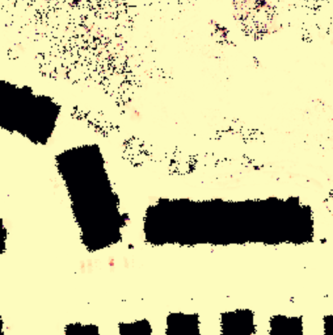} &
\img{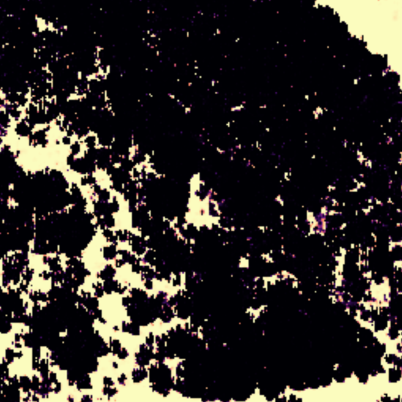} &
\img{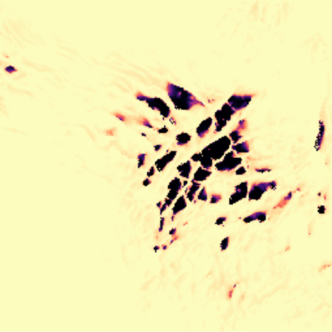} \\

 &
 \barelement{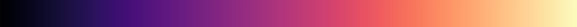}{0.0}{1.0}
 &
\barelement{figures/dtm_generation_results/probs_bar.png}{0.0}{1.0}&
\barelement{figures/dtm_generation_results/probs_bar.png}{0.0}{1.0}\\

\rotatebox{90}{\small DSM} &
\img{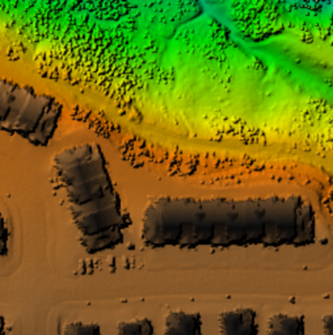} &
\img{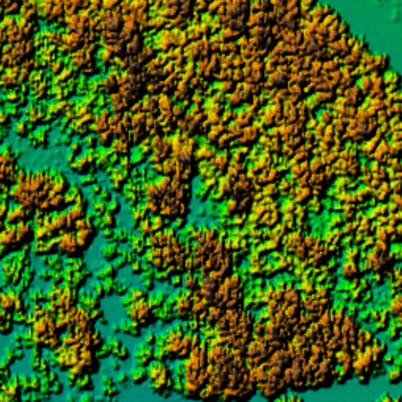} &
\img{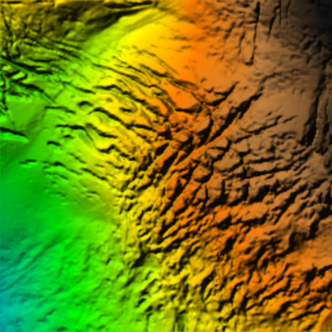} \\

\rotatebox{90}{\small GT DTM} &
\img{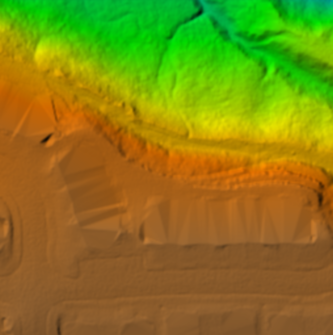} &
\img{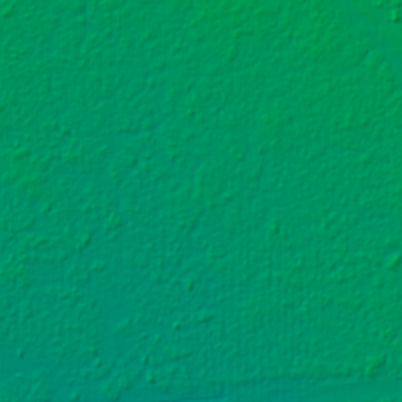} &
\img{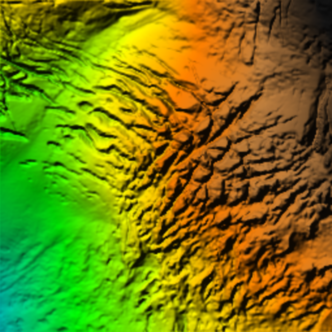} \\

\rotatebox{90}{\small Pred.~DTM} &
\img{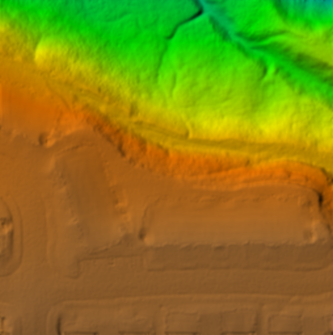} &
\img{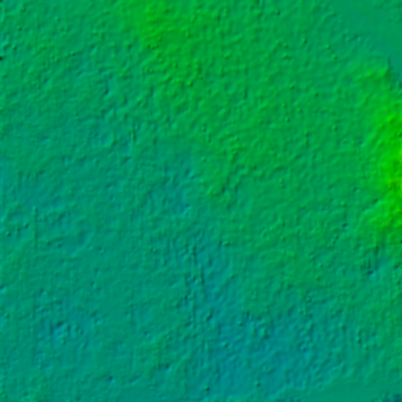} &
\img{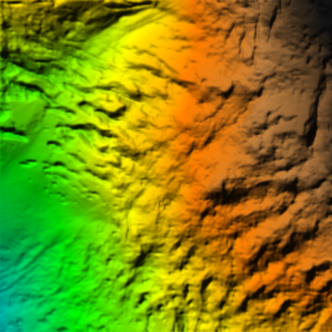} \\

&
\barelement{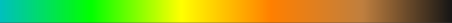}{1410m}{1450m} &
\barelement{figures/dtm_generation_results/elev_bar.png}{12m}{23m} &
\barelement{figures/dtm_generation_results/elev_bar.png}{2150m}{2330m} \\

\rotatebox{90}{\small Rel. Error} &
\img{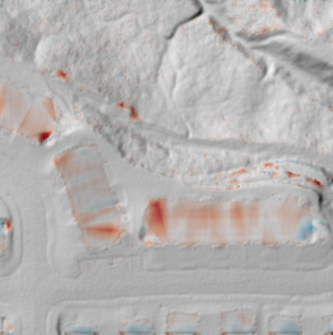} &
\img{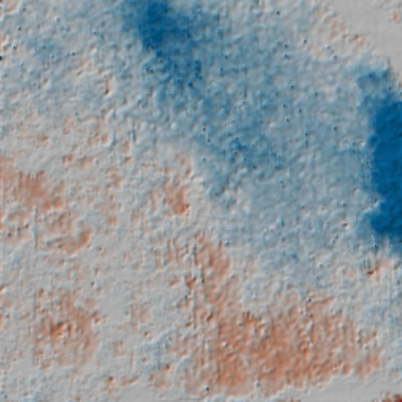} &
\img{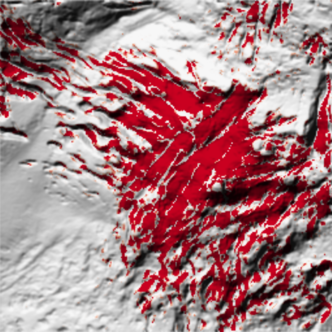} \\

&
\barelement{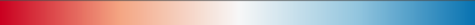}{-1.0m}{1.0m} &
\barelement{figures/dtm_generation_results/error_bar.png}{-1.0m}{1.0m} &
\barelement{figures/dtm_generation_results/error_bar.png}{-1.0m}{1.0m} \\
\end{tabular}
}

\caption{\textbf{Example DTM extraction results across diverse environments in USGS datasets.} Our method effectively removes buildings, vegetation, and other above-ground structures while preserving terrain features. From top to bottom: predicted ground-truth probability maps, input \ac{DSM}, ground-truth \ac{DTM}, our predicted \ac{DTM}, and relative error map.}
\label{fig:regression_results}
\vspace*{-1.25\baselineskip}
\end{figure}

\paragraph{Qualitative analysis.}
\cref{fig:regression_results} shows representative results of \ac{method} across diverse environments. It reliably removes buildings, vegetation, and other above-ground structures while preserving natural terrain features, producing accurate and finely detailed classification maps.

In suburban regions (SU-II), larger errors often occur where the ground-truth \acp{DTM} were generated by simple triangulation rather than direct measurements. This indicates that our diffusion-based formulation yields more plausible terrain reconstructions than interpolation methods. In tree-dominated regions (RT), interpolation errors are higher than in urban scenes; nevertheless, the recovered terrain remains accurate and visually consistent. For rocky mountain areas (KW), strong elevation gradients are occasionally misclassified as structural elements, reflecting an out-of-distribution challenge where high-gradient features resemble building facades in the training data. Including more diverse mountainous samples would improve performance on such terrain.

Additional examples and detailed analysis are provided in the supplementary material.

\subsection{Road Reconstruction Results}
To assess \ac{method}’s applicability to high-precision and smoothness-critical tasks, such as road surface reconstruction, we evaluate it on the GeRoD dataset~\cite{flexroad}. For this task, we use the model trained on SU-I~\cite{su_2020} without any domain-specific fine-tuning, testing cross-regional generalization to an entirely different geographic area and application domain.

\begin{table}[ht]
\centering

\footnotesize
\begin{tabular}{lcccc}
\toprule
\multirow{2}{*}{\bfseries{Method}} & \multicolumn{2}{c}{\bfseries{Road}} & \multicolumn{2}{c}{\bfseries{Non-Road}} \\
& \bfseries{MED}$\downarrow$ & \bfseries{MAD}$\downarrow$ & \bfseries{MED}$\downarrow$ & \bfseries{MAD}$\downarrow$ \\
\midrule
Plane & 2.050 & \textbf{0.000} & 2.343 & \textbf{0.000} \\
\ac{RGT} & 0.415 & 15.550 & 1.773 & 22.191 \\
FlexRoad~\cite{flexroad} & 0.483 & 1.010 & 0.332 & 3.019 \\
\midrule
\ac{method}~{\scriptsize (Ours)} & \textbf{0.078} & 1.492 & \underline{0.123} & 1.801 \\
\ac{method}+~{\scriptsize (Ours)} & \underline{0.109} & \underline{0.626} & \textbf{0.139} & \underline{0.434} \\
\bottomrule
\end{tabular}

\caption{\textbf{Quantitative evaluation on GeRoD dataset~\cite{flexroad}.} Road surface reconstruction performance comparing \ac{MED} in meters and \ac{MAD} in degrees (roughness, lower is better) across road and non-road regions. \ac{method} achieves best accuracy while maintaining competitive smoothness. \ac{method}+ additionally optimizes geometric smoothness. \textbf{Bold}: best performance, \underline{underlined}: second best.}
\label{tab:gerod_results}
\vspace*{-1.25\baselineskip}
\end{table}

\cref{tab:gerod_results} presents quantitative results comparing \ac{method} to specialized road reconstruction methods. Our approach reduces \ac{MED} by 81\% and 63\% for road and terrain regions, respectively, outperforming all baselines. While the Plane method achieves perfect smoothness (\ac{MAD}=0) by definition, it sacrifices accuracy. \ac{method} maintains competitive smoothness (only 48\% higher \ac{MAD} than FlexRoad for roads) while dramatically improving accuracy.

Our enhanced variant, \ac{method}+, applies additional Laplacian smoothing (20 iterations with a smoothing factor of 0.5), which significantly reduces surface roughness. Specifically, it lowers \ac{MAD} by approximately 38\% on road regions and 86\% on non-road regions, while maintaining highest accuracy (\ac{MED} remains comparable to the base \ac{method}). This demonstrates that minor post-processing can effectively optimize the smoothness-accuracy trade-off.

\begin{figure}[ht]
\centering
\vspace{0.5\baselineskip}
\resizebox{\linewidth}{!}{%
\setlength{\tabcolsep}{1pt}       
\renewcommand{\arraystretch}{0.5} 
\begin{tabular}{@{} >{\centering\arraybackslash}m{0.55cm} *{3}{>{\centering\arraybackslash}m{0.28\linewidth}} @{}}
& {\small DOP} & {\small DSM Mesh} & {\small GT DTM Mesh} \\ 
\cmidrule(lr){2-4}

 &
\ImgWithValue[]{\linewidth}{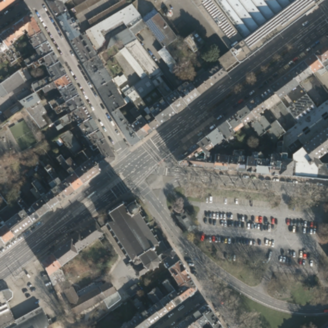} &
\includegraphics[width=\linewidth]{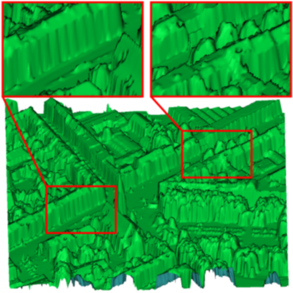} &
\includegraphics[width=\linewidth]{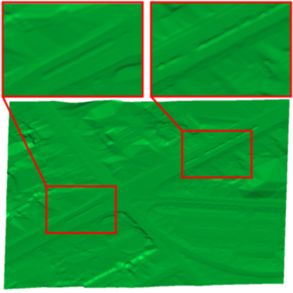} \\

& {\small MED} & {\small MAD} & {\small Ground Mesh} \\ 
\cmidrule(lr){2-4}

\rotatebox{90}{\small Plane} &
\ImgWithValue[1.72m / 2.53m]{\linewidth}{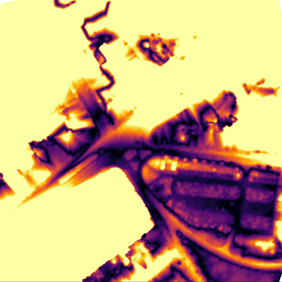} &
\ImgWithValue[0.0° / 0.0°]{\linewidth}{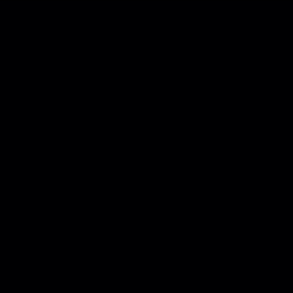} &
\includegraphics[width=\linewidth]{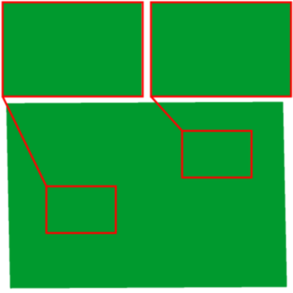} \\

\rotatebox{90}{\small RGT} &
\ImgWithValue[0.77m / 0.34m]{\linewidth}{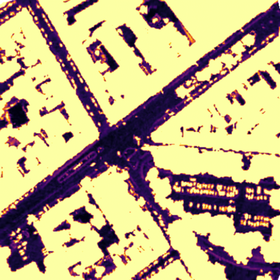} &
\ImgWithValue[18.72° / 7.17°]{\linewidth}{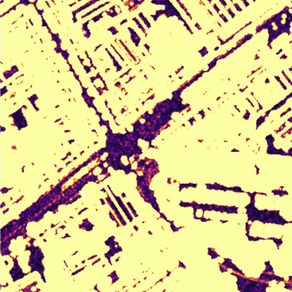} &
\includegraphics[width=\linewidth]{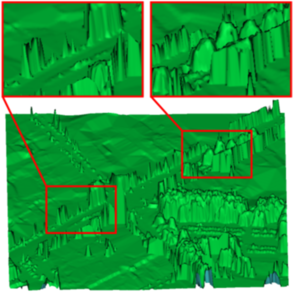} \\

\rotatebox{90}{\small FlexRoad~\cite{flexroad}} &
\ImgWithValue[0.18m / 0.27m]{\linewidth}{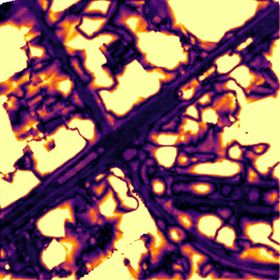} &
\ImgWithValue[0.78° / 2.59°]{\linewidth}{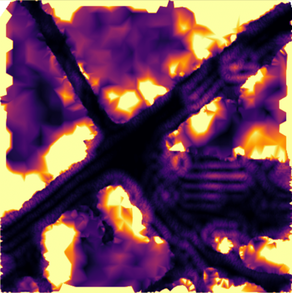} &
\includegraphics[width=\linewidth]{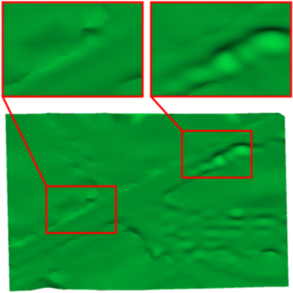} \\

\rotatebox{90}{\small \ac{method}~{\scriptsize (Ours)}} &
\ImgWithValue[0.13m / 0.29m]{\linewidth}{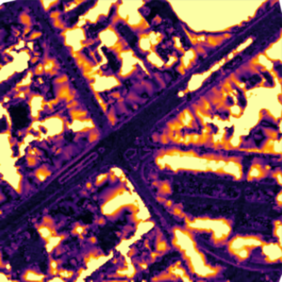} &
\ImgWithValue[2.8° / 4.5°]{\linewidth}{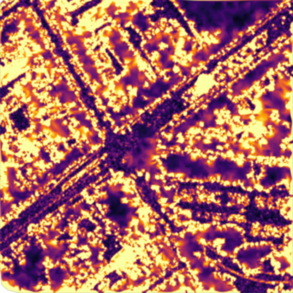} &
\includegraphics[width=\linewidth]{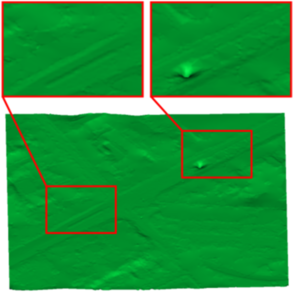} \\

\rotatebox{90}{\small \ac{method}~${}^\dagger$~{\scriptsize (Ours)}} &
\ImgWithValue[0.11m / 0.27m]{\linewidth}{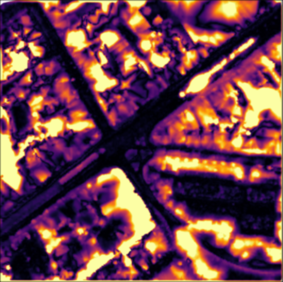} &
\ImgWithValue[0.47° / 1.16°]{\linewidth}{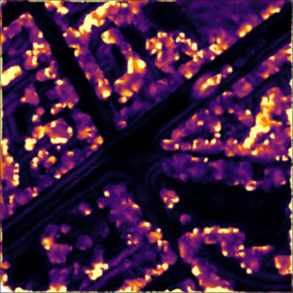} &
\includegraphics[width=\linewidth]{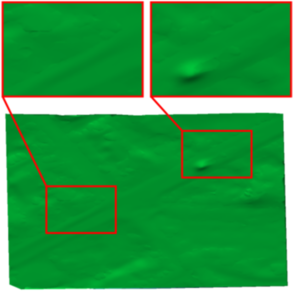} \\
&
\barelement{figures/dtm_generation_results/probs_bar.png}{0m}{0.5m} &
\barelement{figures/dtm_generation_results/probs_bar.png}{0°}{5°} & \\

\end{tabular}%
}
\caption{\textbf{Comparison of road reconstruction methods on the GeRoD dataset~\cite{flexroad}.} Columns show \ac{MED}, \ac{MAD}, and 3D mesh (dynamic sampling~\cite{flexroad} was also used). Rows correspond to the evaluated methods. Metric values indicate the mean of the error map for road / terrain regions.}
\label{fig:road}
\vspace*{-1.25\baselineskip}
\end{figure}

\cref{fig:road} demonstrates that, despite this general formulation, our method produces accurate, fine-grained road surfaces. Applying a simple smoothing post-processing step improves road roughness and makes the surface more geometrically accurate, but only minimally decreases precision, indicating no strong trade-off between precision and smoothness. The figure also shows an artifact that appears in all reconstructions due to a very noisy region in the DSM; our method exhibits the least pronounced effect. Notably, without fine-tuning on these unseen regions, the strong cross-regional performance indicates that our diffusion-based formulation captures fundamental terrain priors that transfer effectively across diverse environments and applications. Additional road reconstruction results are in the supplementary.

\subsection{Ablation Studies}
\label{sec:ablation}

We conduct ablation studies on the GeRoD dataset~\cite{flexroad} to analyze the impact of key design choices. The dataset is split into train/validation/test sets, where validation includes a suburban region (381\_*), and test comprises a rural highway region (407\_*) and an urban roundabout area (296\_*).

\begin{table}[ht]
\centering
\resizebox{\linewidth}{!}{%
\setlength{\tabcolsep}{2pt}
\footnotesize
\begin{tabular}{lccccc}
\toprule
\bfseries{Variant} & \bfseries{RMSE}$\downarrow$ & \bfseries{MAE}$\downarrow$ & \bfseries{$\boldsymbol{E_{T_1}}$}$\downarrow$ & \bfseries{$\boldsymbol{E_{T_2}}$}$\downarrow$ & \bfseries{$\boldsymbol{E_{tot}}$}$\downarrow$ \\
\midrule
w/o diffusion process (UNet only) & 1.895 & 1.372 & 38.92 & 8.90 & 35.95 \\
Target: Absolute DTM & 0.842 & 0.427 & 1.62 & 1.08 & 1.20 \\
w/o Gating & 8.937 & 6.062 & 8.58 & 52.72 & 52.64 \\
\midrule
\textbf{Baseline (Ours)} & \textbf{0.700} & \underline{0.393} & 1.43 & 1.06 & 1.11 \\
\bottomrule
\end{tabular}
}
\caption{\textbf{Ablation studies on GeRoD dataset~\cite{flexroad}.} 
Comparison of diffusion vs. single-step UNet, target formulation, and gating mechanism. RMSE and MAE in meters; classification errors as percentages. \textbf{Bold}: best, \underline{underlined}: second best.}
\label{tab:architecture_ablation_main}
\vspace*{-0.25\baselineskip}
\end{table}

Results in \cref{tab:architecture_ablation_main} highlight three key factors.
First, removing the diffusion process and using a single UNet pass (the same architecture as our denoiser) drastically increases errors---especially in classification---showing that iterative refinement is essential for ground filtering.
Second, predicting residuals (\ac{nDSM}) instead of absolute elevations improves accuracy by 17\%, as the network can focus on non-ground structures while using the DSM as a strong prior.
Third, removing the gating mechanism causes severe performance degradation, as it guides the diffusion process by separating terrain from above-ground structures and leveraging the model’s generative capability during interpolation.
Overall, diffusion yields a 63\% RMSE reduction over single-step prediction, residual learning reduces errors by 17\%, and removing gating causes a 12× drop in performance. More ablations are provided in the supplementary.
\section{Conclusion}
We introduced \ac{method}, a diffusion-based framework for generating \acp{DTM} from \acp{DSM}. It combines a conditional diffusion process that treats non-ground structures as noise, a gated denoiser with confidence-guided reconstruction, and \ac{stitching} for large-scale modeling. Across multiple benchmarks, \ac{method} outperforms state-of-the-art methods, reducing RMSE by up to 93\% on ALS2DTM and 47\% on USGS. For road reconstruction, it achieves 81\% lower mean Euclidean distance than FlexRoad, while \ac{method}+ improves smoothness by 38\% on roads and 85\% on non-road regions.
\paragraph{Limitations and future work.} Although \ac{method} generalizes well, it can struggle with out-of-distribution terrain with abrupt elevation changes. Future work may explore point-based diffusion for direct LiDAR processing and improve robustness through training on more diverse terrain.
{
    \small
    \bibliographystyle{ieeenat_fullname}
    \bibliography{main}
}

\clearpage
\setcounter{page}{1}
\maketitlesupplementary

\noindent
This supplementary material provides additional implementation details, extended ablation studies, and qualitative results to complement the main paper. We organize the content as follows:

\begin{itemize}
    \item \cref{sec:supp_datasets}: Dataset Details 
    \item \cref{sec:supp_implementation}: Implementation Details
    \item \cref{sec:supp_hardware}: Hardware and Timing Performance
        \item \cref{sec:supp_diffusion}: Analysis of Diffusion Steps
    \item \cref{sec:supp_qualitative}: Additional Qualitative Results
    \item \cref{sec:supp_groundiff}: Ablations on \ac{method}
    \item \cref{sec:supp_stitching}: Ablations on \ac{stitching}
    \item \cref{sec:supp_limitations}: Limitations and Failure Cases
    \item \cref{sec:supp_ethic}: Ethical Considerations
\end{itemize}

\section{Dataset Details}
\label{sec:supp_datasets}

\subsection{ALS2DTM Datasets}
\begin{figure*}[ht]
\centering
\setlength{\tabcolsep}{1pt}       
\renewcommand{\arraystretch}{0.6} 

\resizebox{\linewidth}{!}{%
\begin{tabular}{@{} >{\centering\arraybackslash}m{0.04\linewidth}
*{7}{>{\centering\arraybackslash}m{0.125\linewidth}} @{}}

& \multicolumn{7}{c}{\textbf{DALES Dataset}} \\ 
\cmidrule(lr){2-8}

& \multicolumn{3}{c}{\textbf{Train}} & \multicolumn{2}{c}{\textbf{Val}} & \multicolumn{2}{c}{\textbf{Test}} \\ 
\cmidrule(lr){2-4} \cmidrule(lr){5-6} \cmidrule(lr){7-8}

\rotatebox{90}{\small DSM} &
\img{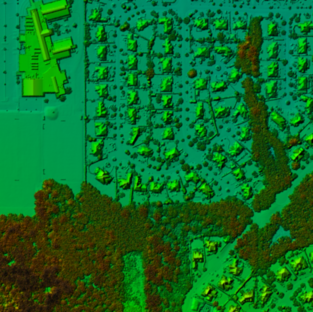} &
\img{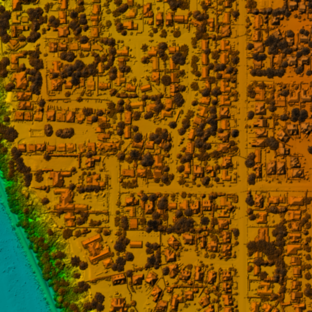} &
\img{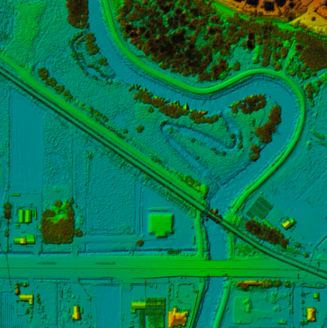} &
\img{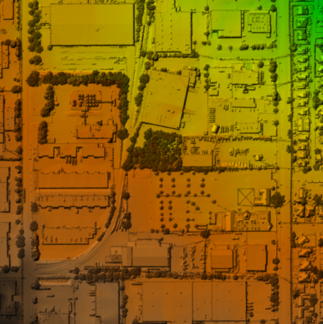} &
\img{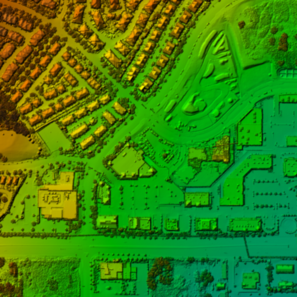} &
\img{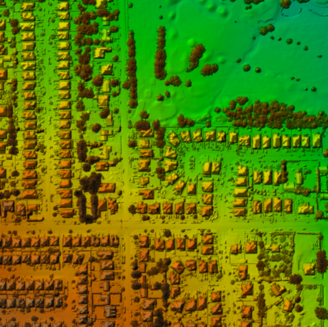} &
\img{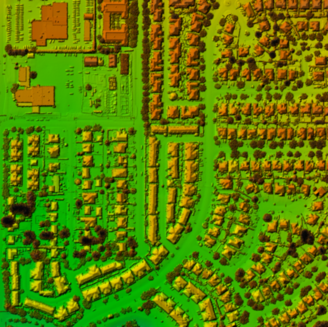} \\

\rotatebox{90}{\small DTM} &
\img{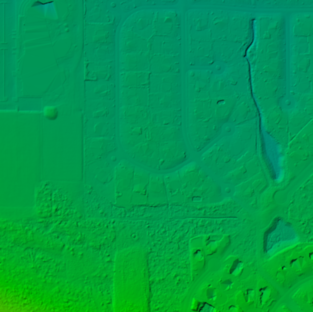} &
\img{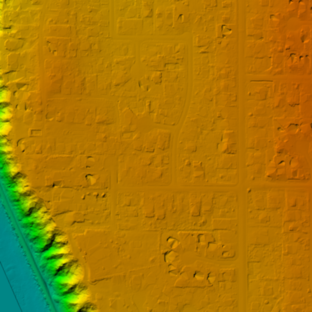} &
\img{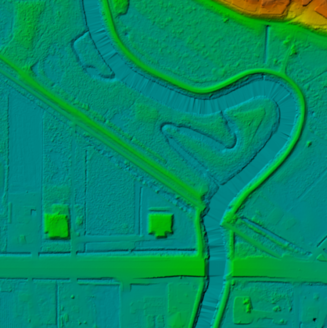} &
\img{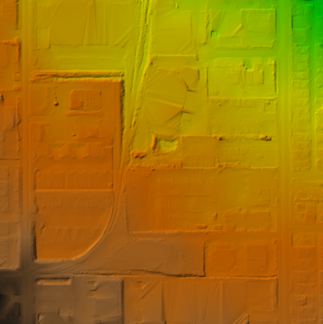} &
\img{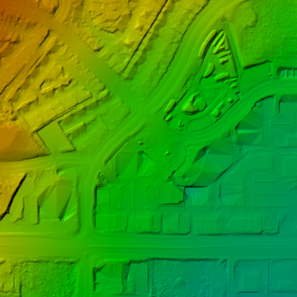} &
\img{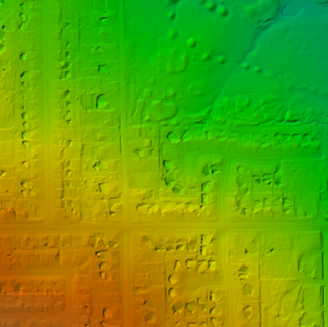} &
\img{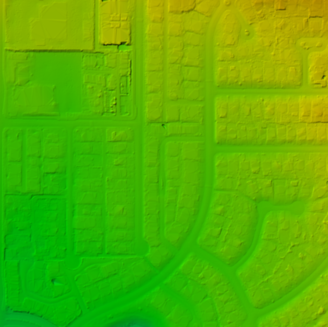} \\

& 
\barelement{figures/dtm_generation_results/elev_bar.png}{80m}{140m} &
\barelement{figures/dtm_generation_results/elev_bar.png}{0m}{80m} &
\barelement{figures/dtm_generation_results/elev_bar.png}{0m}{20m} &
\barelement{figures/dtm_generation_results/elev_bar.png}{50m}{75m} &
\barelement{figures/dtm_generation_results/elev_bar.png}{10m}{80m} &
\barelement{figures/dtm_generation_results/elev_bar.png}{10m}{60m} &
\barelement{figures/dtm_generation_results/elev_bar.png}{60m}{90m} \\

\midrule

& \multicolumn{7}{c}{\textbf{NB Dataset}} \\ 
\cmidrule(lr){2-8}

& \multicolumn{3}{c}{\textbf{Train}} & \multicolumn{2}{c}{\textbf{Val}} & \multicolumn{2}{c}{\textbf{Test}} \\ 
\cmidrule(lr){2-4} \cmidrule(lr){5-6} \cmidrule(lr){7-8}

\rotatebox{90}{\small DSM} &
\img{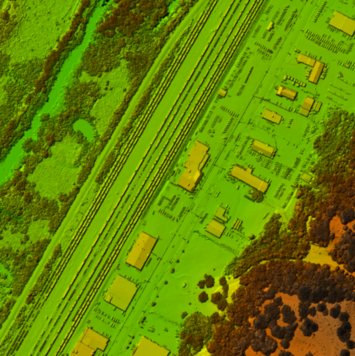} &
\img{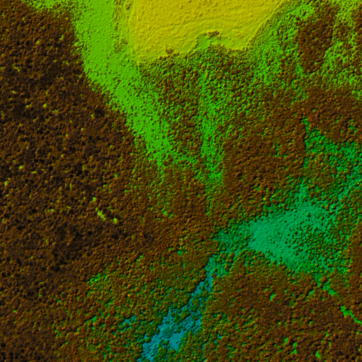} &
\img{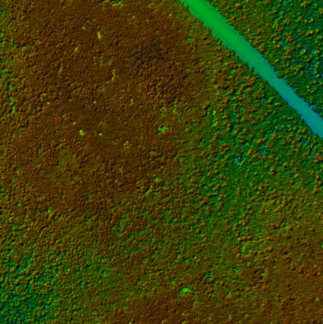} &
\img{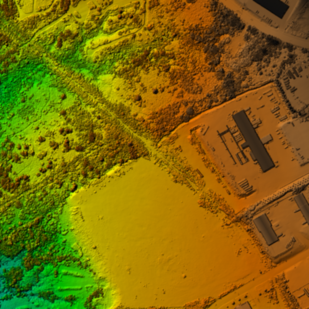} &
\img{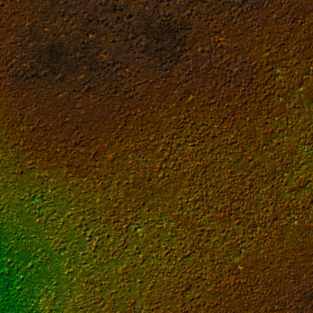} &
\img{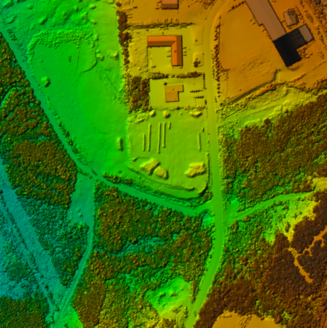} &
\img{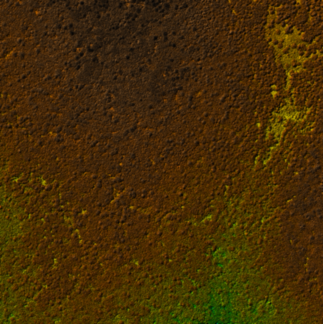} \\

\rotatebox{90}{\small DTM} &
\img{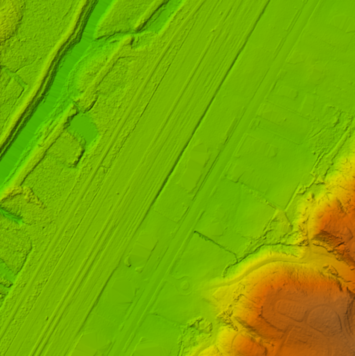} &
\img{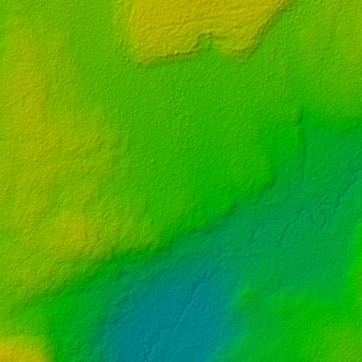} &
\img{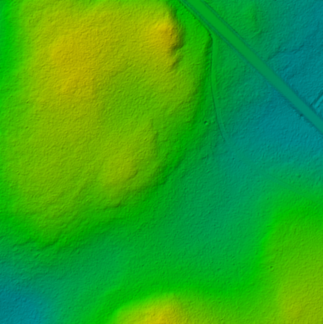} &
\img{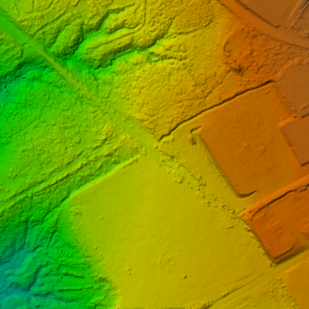} &
\img{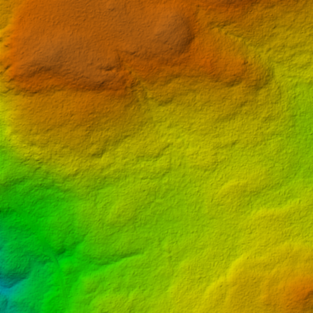} &
\img{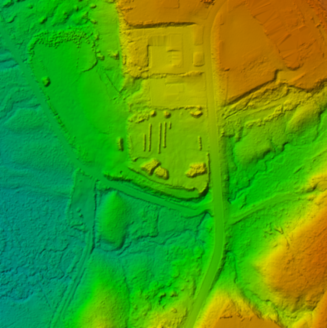} &
\img{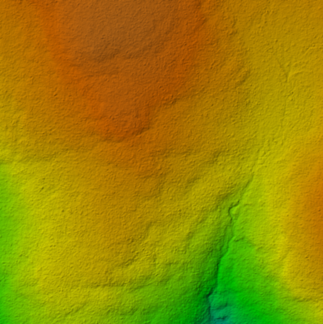} \\

& 
\barelement{figures/dtm_generation_results/elev_bar.png}{-20m}{50m} &
\barelement{figures/dtm_generation_results/elev_bar.png}{335m}{360m} &
\barelement{figures/dtm_generation_results/elev_bar.png}{330m}{370m} &
\barelement{figures/dtm_generation_results/elev_bar.png}{5m}{50m} &
\barelement{figures/dtm_generation_results/elev_bar.png}{230m}{300m} &
\barelement{figures/dtm_generation_results/elev_bar.png}{3m}{50m} &
\barelement{figures/dtm_generation_results/elev_bar.png}{370m}{330m}\\

\end{tabular}%
}

\caption{\textbf{Representative terrain types from the ALS2DTM benchmark datasets~\cite{le2022learning}}. Each block shows DSM (top), corresponding DTM (middle), and elevation bars (bottom) for training, validation, and test splits. Top block: DALES dataset; bottom block: NB dataset.}
\label{fig:als2dtm_samples}
\vspace*{-1\baselineskip}
\end{figure*}

The ALS2DTM benchmark datasets~\cite{le2022learning} consist of \textbf{DALES} and \textbf{NB}, both with predefined train/validation/test splits. DALES contains 29 training, 10 validation, and 11 test samples, while the NB dataset comprises 84 training, 42 validation, and 42 test samples, each covering a $500\,\text{m} \times 500\,\text{m}$ area at $0.1\,\text{m/px}$ resolution. The \acp{DSM} were generated via maximum grid sampling from LiDAR data, while the \acp{DTM} were obtained from previous work~\cite{le2022learning}: the DALES \acp{DTM} were acquired from the Canadian governmental geoportal, whereas the NB \acp{DTM} were produced through ground classification, manual corrections, and interpolation using \ac{TIN}. The NB dataset includes a variety of topologies, ranging from urban and suburban areas to forests, whereas DALES is limited to urban scenes. Representative examples of samples from both datasets are shown in~\cref{fig:als2dtm_samples}.

\subsection{USGS (OpenTopology) Datasets}

We utilize three regions from the OpenTopology portal, following prior work:  

\paragraph{SU (Salt Lake City, Utah).} Captured over 2013–2014 using airborne LiDAR, this region covers all of Salt Lake City, totaling approximately $1360~\text{km}^2$. Due to its large size, SU was divided into three datasets in previous work~\cite{amirkolaee2022dtm}: SU-I ($\sim 7521~\text{m} \times 3871~\text{m}$), SU-II ($\sim 7090~\text{m} \times 3640~\text{m}$), and SU-III ($\sim 6718~\text{m} \times 3092~\text{m}$), all at 0.5\,m/px resolution. Each dataset covers approximately 90\%, 80\%, and 40\% urban areas, respectively, with the remainder consisting of mountainous terrain. The three datasets are shown in~\cref{fig:su_i_dataset,fig:su_ii_dataset,fig:su_iii_dataset}.

\begin{figure}[ht]
\centering
\setlength{\tabcolsep}{1pt}       
\renewcommand{\arraystretch}{0.6} 

\resizebox{\linewidth}{!}{%
\begin{tabular}{@{} *{2}{>{\centering\arraybackslash}m{0.49\linewidth}} @{}}

DSM & DTM \\ 

\includegraphics[width=\linewidth]{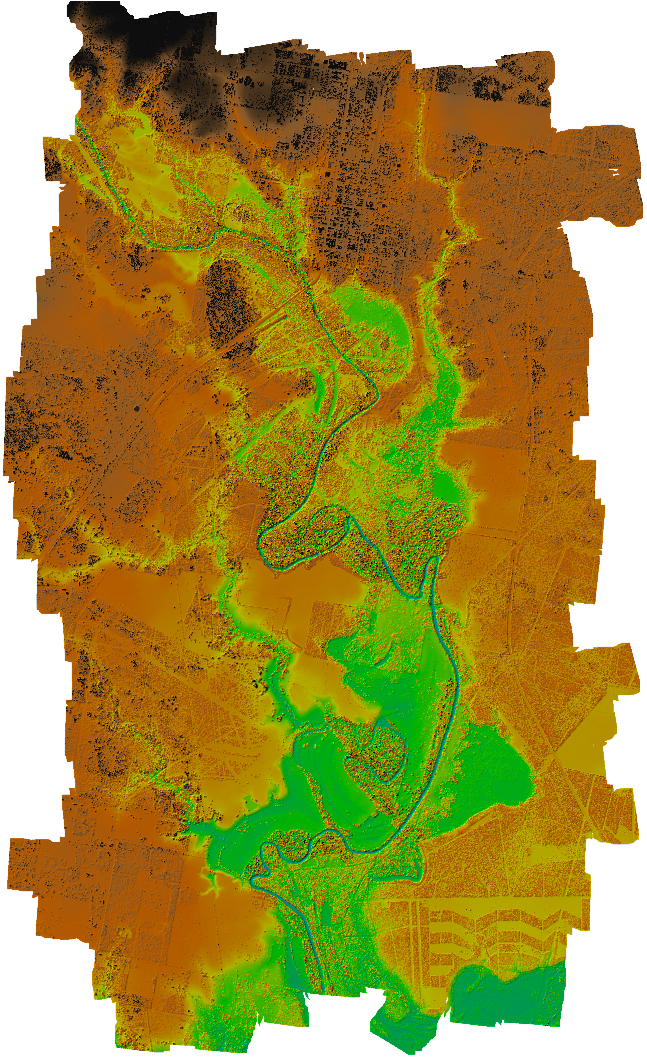} & \includegraphics[width=\linewidth]{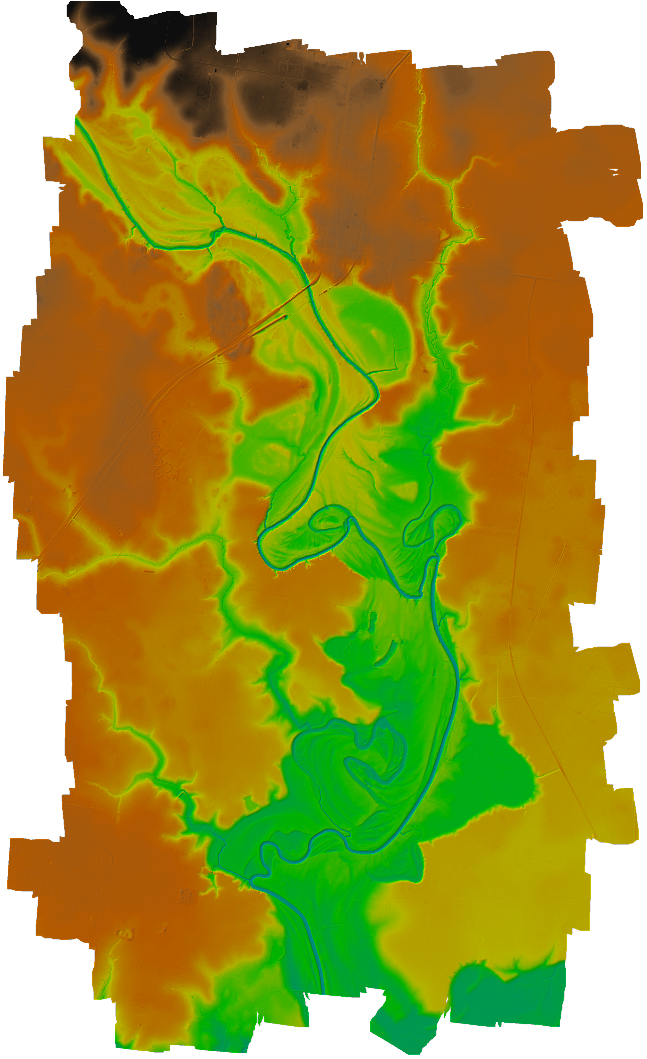} \\

\multicolumn{2}{c}{\barelement{figures/dtm_generation_results/elev_bar.png}{0m}{20m}} \\

\end{tabular}%
}

\caption{\textbf{Visualization of the RT dataset~\cite{rt_2018} visualization}. Left: DSM, right: DTM, bottom: elevation bar.}
\label{fig:rt_dataset}
\vspace*{-1\baselineskip}
\end{figure}

\paragraph{RT (Refugio, Texas).} Acquired by the National Center for Airborne Laser Mapping (NCALM) along the Mission River in Refugio, Texas, following Hurricane Harvey on August 5–6, 2018, using airborne LiDAR, this dataset spans $7196~\text{m} \times 11883~\text{m}$ at 1\,m/px resolution. It covers approximately 90\% rural area, including a river and plantation regions (agricultural fields or forested areas) with varying vegetation height, while the remaining 10\% is suburban. The region is illustrated in~\cref{fig:rt_dataset}.

\begin{figure}[ht]
\centering
\setlength{\tabcolsep}{1pt}       
\renewcommand{\arraystretch}{0.6} 

\resizebox{\linewidth}{!}{%
\begin{tabular}{@{} *{2}{>{\centering\arraybackslash}m{0.49\linewidth}} @{}}

DSM & DTM \\ 

\includegraphics[width=\linewidth]{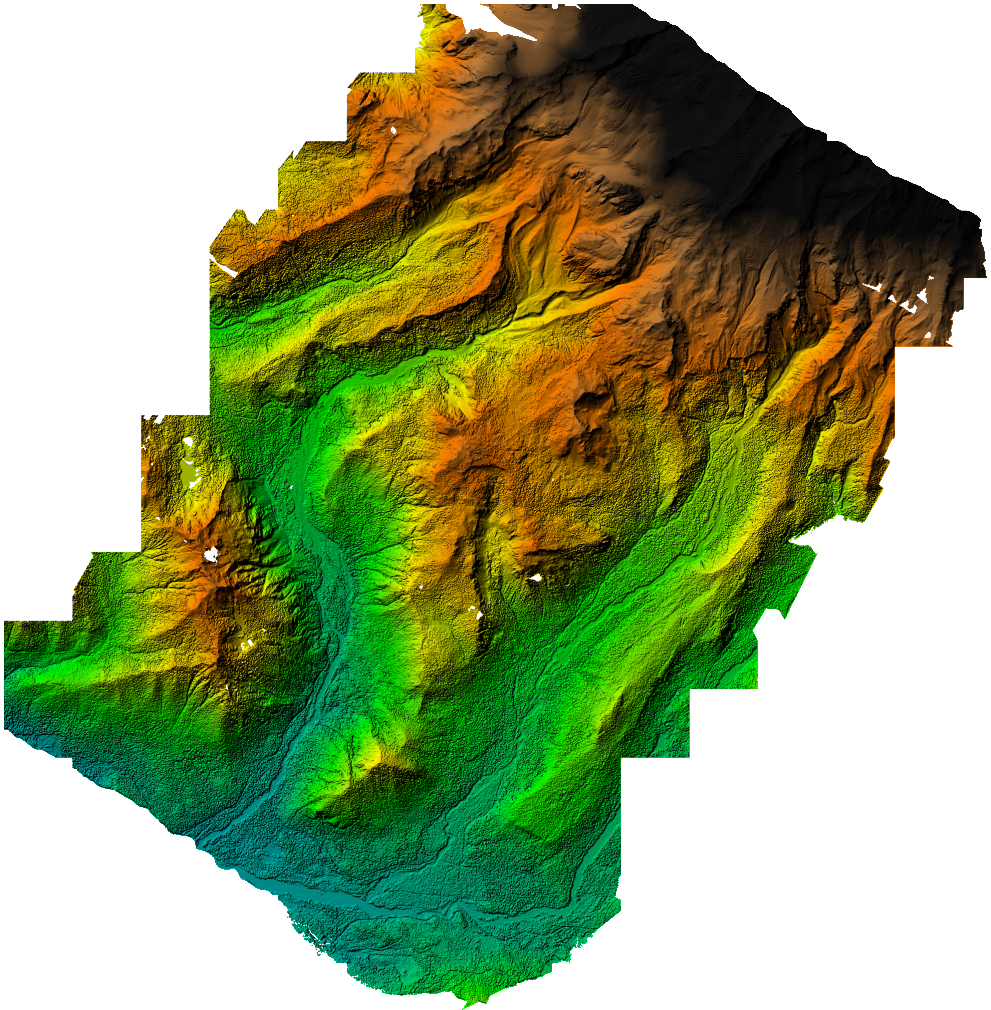} & \includegraphics[width=\linewidth]{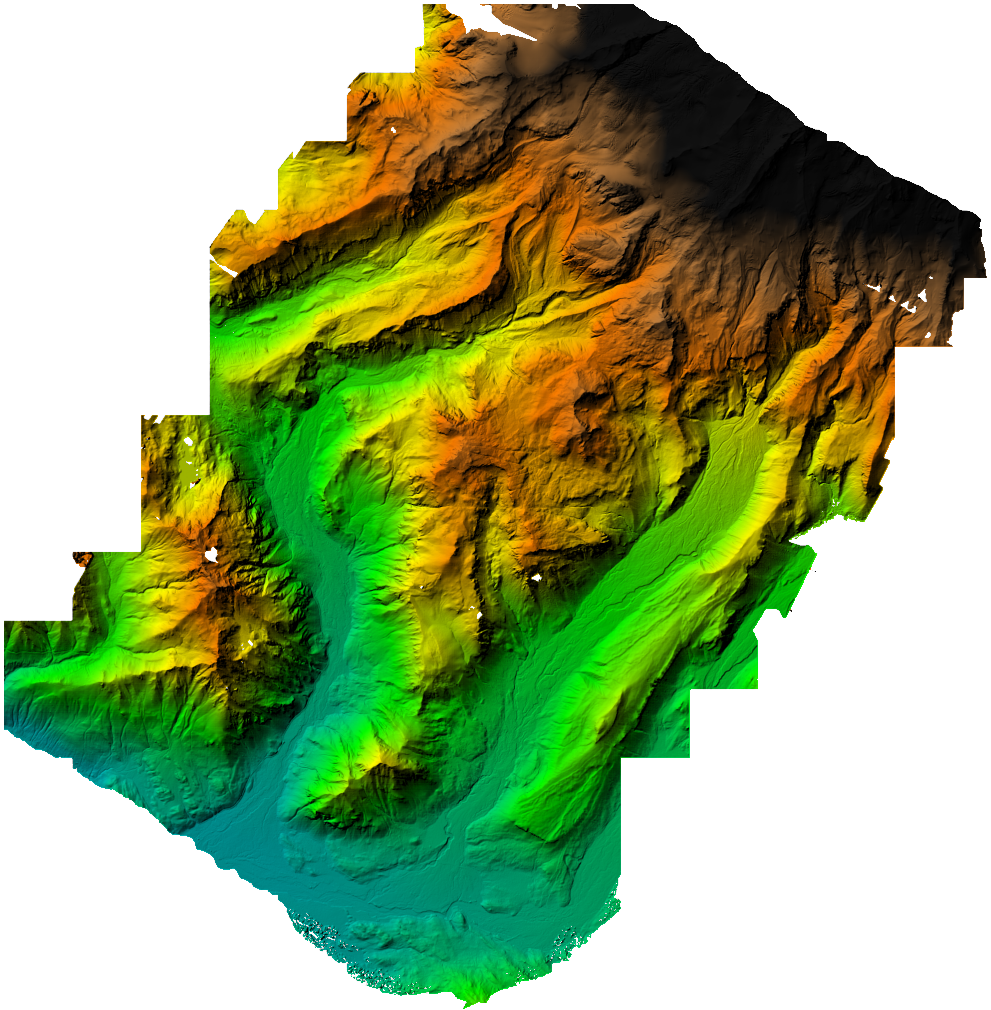} \\

\multicolumn{2}{c}{\barelement{figures/dtm_generation_results/elev_bar.png}{650m}{2400m}} \\

\end{tabular}%
}

\caption{\textbf{Visualization of the KW dataset ~\cite{kw_2020}}. Left: DSM, right: DTM, bottom: elevation bar.}
\label{fig:kw_dataset}
\vspace*{-1\baselineskip}
\end{figure}

\paragraph{KW (Kautz Creek, Washington).} Captured on August 28, 2012, within Mount Rainier National Park, Washington, this dataset covers the Kautz Creek watershed ($581{,}000~\text{m} \times 5{,}189{,}000~\text{m}$) at 1\,m/px resolution. It was collected to study landscape response to debris flows and associated hazards. The area features steep mountainous terrain with abrupt elevation changes (alpine) and low-growing vegetation. \cref{fig:kw_dataset} provides a visualization of the dataset.
\newline
\newline
For consistency with our \ac{method}, all datasets are divided into $256 \times 256$ patches, resulting in 1734 SU-I, 1540 SU-II, 1247 SU-III, 4018 RT, and 1760 KW samples. All \acp{DSM} and \acp{DTM} were downloaded from the OpenTopography portal.


\begin{figure}[ht]
\centering
\setlength{\tabcolsep}{1pt}       
\renewcommand{\arraystretch}{0.6} 

\resizebox{\linewidth}{!}{%
\begin{tabular}{@{} >{\centering\arraybackslash}m{0.04\linewidth}
*{1}{>{\centering\arraybackslash}m{0.875\linewidth}} @{}}

\rotatebox{90}{\small DSM} &
\img{figures/datasets/su_I_dsm.png} \\

\rotatebox{90}{\small DTM} &
\img{figures/datasets/su_I_dtm.png} \\
& 
\barelement{figures/dtm_generation_results/elev_bar.png}{1350m}{1800m} \\

\end{tabular}%
}

\caption{\textbf{Visualization of the SU-I dataset~\cite{su_2020}}. Top: DSM, middle: DTM, bottom: elevation bar.}
\label{fig:su_i_dataset}
\end{figure}

\begin{figure}[ht]
\centering
\setlength{\tabcolsep}{1pt}       
\renewcommand{\arraystretch}{0.6} 

\resizebox{\linewidth}{!}{%
\begin{tabular}{@{} >{\centering\arraybackslash}m{0.04\linewidth}
*{1}{>{\centering\arraybackslash}m{0.875\linewidth}} @{}}

\rotatebox{90}{\small DSM} &
\img{figures/datasets/su_II_dsm.png} \\

\rotatebox{90}{\small DTM} &
\img{figures/datasets/su_II_dtm.png} \\
& 
\barelement{figures/dtm_generation_results/elev_bar.png}{1350m}{2200m} \\

\end{tabular}%
}

\caption{\textbf{Visualization of the SU-II dataset~\cite{su_2020}}. Top: DSM, middle: DTM, bottom: elevation bar.}
\label{fig:su_ii_dataset}
\vspace*{-1\baselineskip}
\end{figure}

\begin{figure}[ht]
\centering
\setlength{\tabcolsep}{1pt}       
\renewcommand{\arraystretch}{0.6} 

\resizebox{\linewidth}{!}{%
\begin{tabular}{@{} >{\centering\arraybackslash}m{0.04\linewidth}
*{1}{>{\centering\arraybackslash}m{0.875\linewidth}} @{}}

\rotatebox{90}{\small DSM} &
\img{figures/datasets/su_III_dsm.png} \\

\rotatebox{90}{\small DTM} &
\img{figures/datasets/su_III_dtm.png} \\
& 
\barelement{figures/dtm_generation_results/elev_bar.png}{1350m}{2500m} \\

\end{tabular}%
}

\caption{\textbf{Visualization of the SU-III dataset~\cite{su_2020}}. Top: DSM, middle: DTM, bottom: elevation bar.}
\label{fig:su_iii_dataset}
\vspace*{0\baselineskip}
\end{figure}

\section{Implementation Details}
\label{sec:supp_implementation}

\subsection{Data Preprocessing}
For datasets providing only LiDAR point clouds (DALES and NB), we generate \acp{DSM} through rasterization by selecting the maximum elevation within each grid cell.

We divide training and validation sets into 256×256 tiles. Our augmentation pipeline includes:
\begin{itemize}
    \item Random rotations from \{0°, 90°, 180°, 270°\} with additional jittering within the (-5°, 5°) range.
    \item Multi-scale resizing to \{256×256, 512×512, 1024×1024\} to simulate varying metric pixel resolutions.
    \item Random cropping of a 256×256 tile.
    \item Horizontal and vertical flipping following~\cite{amirkolaee2022dtm}.
\end{itemize}
Each augmentation step is applied with 0.5 probability. Augmentation is followed by resizing to 256×256 to match network input requirements.

\subsection{Normalization Strategy}
We employ min-max normalization computed from valid pixels across both \ac{DSM} and \ac{DTM}, mapping all values to the [-1, 1] range while setting invalid regions to zero. Specifically, we calculate the global minimum from both rasters' minima and the global maximum from both rasters' maxima, then apply the transformation:
\begin{align}
x_{\text{norm}} =
2 \cdot \frac{x - \min(s_m, g_m)}{\max(s_m, g_m) - \min(s_m, g_m)} - 1\, ,
\end{align}
where $s_m$ and $g_m$ denote the sets of valid pixels in the \ac{DSM} $s$ and the \ac{DTM} $g$ respectively, as defined by the mask $m$.
This approach contrasts with prior methods using global standardization~\cite{bittner2023dsm2dtm} or data localization~\cite{amirkolaee2022dtm}, providing better numerical stability for diffusion processes, as demonstrated in our ablation study. Binary masks $m$ indicating valid pixels undergo identical augmentation transformations to ensure spatial consistency and exclude invalid regions from loss computation.

\subsection{Training Configuration}
Networks are trained using the AdamW optimizer~\cite{loshchilov2018decoupled} with learning rate 1e-4, weight decay 0.01, and maximum 1000 epochs with early stopping. A cosine annealing scheduler with 500 warmup steps controls learning rate decay. The diffusion process uses $T=10$ denoising steps by default unless otherwise specified, with a cosine noise scheduler ranging from 0.0001 to 0.02. Training uses batches of size 16. Loss hyperparameters are empirically set as: $\lambda_1 = 1.0$, $\lambda_2 = 1.0$, $\lambda_\nabla = 0.1$, $\lambda_c = 0.1$.

\section{Hardware and Timing Performance}
\label{sec:supp_hardware}

\subsection{Hardware Requirements}
Our \ac{method} model contains 62.6M parameters and requires approximately 500MB of memory during inference. All training and testing are conducted on NVIDIA A40 GPUs with 48GB VRAM using PyTorch.

\subsection{Training Time}
Training a single model takes approximately 6 hours to 1 day on an NVIDIA A40 GPU, depending on the dataset and experiment configuration. Larger timestep values require more time as validation steps are slower. Convergence typically occurs within 10K to 20K iterations depending on dataset complexity.

\subsection{Inference Speed}
During inference, a single reverse diffusion step on a $256\times256$ tile takes approximately $60$\,ms on our GPU. 
For $T$ diffusion steps, the per-tile inference time is:
\begin{align}
t_{\text{tile}} = 0.06 \cdot T \quad \text{[s]}.
\end{align}
Given an input of width $W$ and height $H$ (in pixels), tile size $P=256$, and stride $S$, 
the number of tiles along each axis is:
\begin{align}
N_x = \left\lceil \frac{W - P}{S} \right\rceil + 1, \qquad
N_y = \left\lceil \frac{H - P}{S} \right\rceil + 1,
\end{align}
yielding a total of:
\begin{align}
N_{\text{tiles}} = N_x \cdot N_y.
\end{align}
The overall inference time becomes:
\begin{align}
t_{\text{total}} = N_{\text{tiles}} \cdot t_{\text{tile}}.
\end{align}

For an area of size $A$ (in km$^2$) at ground sampling distance $r$ (in m/pixel), 
the image side length (in pixels) is:
\begin{align}
W = H = \frac{1000 \cdot \sqrt{A}}{r},
\end{align}
which directly determines $N_{\text{tiles}}$ through the equations above. 
The scaling is approximately linear with area and quadratic with resolution. We show approximate timing examples in~\cref{tab:processing_time}.

\begin{table}[ht]
\centering
\begin{tabular}{cccc}
\toprule
\textbf{Area} & \textbf{Resolution} & \textbf{Stride} & \textbf{Time} \\
\midrule
1 km$^2$  & 1.0 & 256 & 2  \\
1 km$^2$  & 0.5 & 256 & 8  \\
1 km$^2$  & 0.5 & 128 & 32 \\
5 km$^2$  & 1.0 & 256 & 10 \\
5 km$^2$  & 1.0 & 128 & 40 \\
10 km$^2$ & 1.0 & 256 & 20 \\
\bottomrule
\end{tabular}
\caption{\textbf{Processing time for different area sizes, spatial resolutions, and tile strides using our \ac{method} with \ac{stitching}.}  
All times are in minutes; resolution is in meters per pixel. All times are reported for $T=10$.}
\label{tab:processing_time}
\end{table}

Compared to a simple divide-and-predict strategy, our \ac{stitching} strategy introduces minimal overhead: the prior \ac{DTM} is computed once from a low-resolution version of the input, and blending operations are negligible as they involve straightforward functions. The dominant computational cost arises from per-tile inference, which scales predictably with the number of tiles and thereby with input size and resolution. Note that we do not include batching or multiprocessing strategies in these timing computations, using only single-tile batches during network inference.

\section{Analysis of Diffusion Steps}
\label{sec:supp_diffusion}

\begin{figure}[ht]
    \centering
    \includegraphics[width=\linewidth]{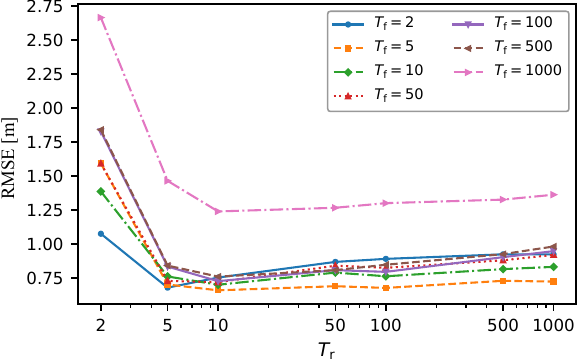}
    \caption{\textbf{RMSE performance across different diffusion timesteps.} 
Models trained with different $T_{\text{f}}$ values are evaluated across varying $T_{\text{r}}$ during testing.}
    \label{fig:ablation_steps}
\end{figure}

We analyze the impact of diffusion steps on model performance through comprehensive experiments on GeRoD splits defined in the main paper. We evaluate diffusion models trained with different forward timesteps ($T_f$) across various reverse inference timesteps ($T_r$). The results in~\cref{fig:ablation_steps} demonstrate that optimal configurations lie in the moderate timestep range.

Models trained with minimal timesteps ($T_f = 2$) exhibit high instability and poor performance across most inference settings, with RMSE increasing when reverse timesteps exceed the training value. Training with only two timesteps is insufficient for denoising, as the setup approaches a single-pass UNet. Conversely, models trained with extensive timesteps ($T_f = 100, 500, 1000$) suffer from degraded performance and prohibitive computational costs, with $T_f = 1000$ producing extremely high RMSE. We hypothesize that this occurs because the denoiser is expected to handle finer-grained structures, which requires higher network capacity and is inherently more challenging.

Both moderate settings ($T_f=5, 10$) are stable and show a performance plateau after reaching the training timesteps, indicating that at least this number of steps is needed to reach optimum performance. We adopt $T_f=10$ as our default configuration because it achieves its lowest RMSE precisely at $T_r=T_f=10$, providing consistency between training and inference regimes while maintaining computational efficiency.

\begin{figure*}[ht]
    \centering
    \includegraphics[width=\linewidth]{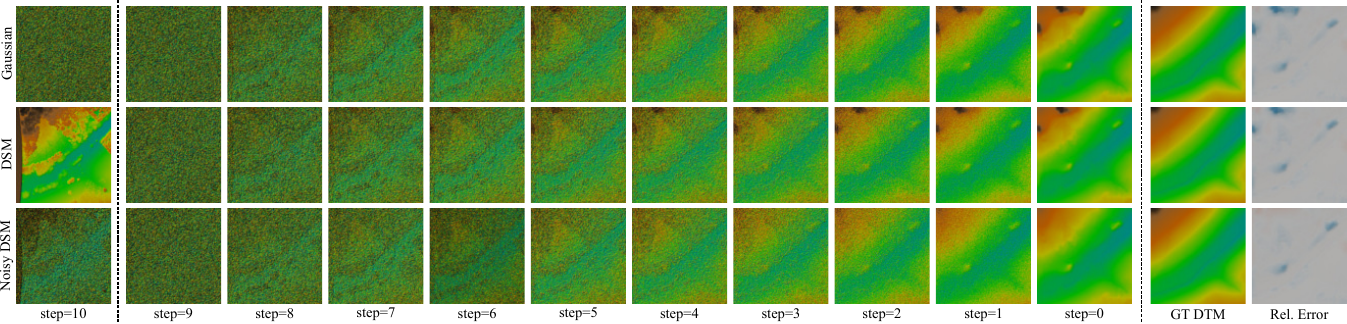} \\[0.5em]
    \vspace{-0.5\baselineskip}

    \resizebox{\linewidth}{!}{%
    \renewcommand{\arraystretch}{0.6}%
    \begin{tabular}{@{}m{0.86\linewidth}%
                    >{\centering\arraybackslash}m{0.08\linewidth}@{}%
                    >{\centering\arraybackslash}m{0.08\linewidth}@{}}
        & \barelement{figures/dtm_generation_results/elev_bar.png}{-1}{+1}%
        &\barelement{figures/dtm_generation_results/error_bar.png}{-1}{+1}
    \end{tabular}%
    }

    \vspace{-0.5\baselineskip}
    \caption{\textbf{Diffusion-based denoising progression for $T_f=T_r=10$.} 
    Progressive generation of cleaner terrain \acp{DTM} conditioned on the \ac{DSM}, 
    starting from Gaussian noise, raw \ac{DSM}, or noisy \ac{DSM}. 
    We show the pure denoiser output terrain $s-\hat{r}$ without gating for clear visualization, 
    highlighting the learned interpolation capability. Intermediate steps progress 
    from the initial input (step 10) to the final output (step 0). 
    Errors are color-encoded from red (-1) to blue (+1), and all elevations are normalized 
    to the [-1, 1] range.}
    \label{fig:denoising_progression}
    \vspace*{-1\baselineskip}
\end{figure*}

\cref{fig:denoising_progression} visualizes the progressive denoising process, showing how \ac{method} iteratively refines terrain structures for $T_f=T_r=10$. Initialization with pure Gaussian noise or raw \ac{DSM} alone results in higher errors than when fusing the \ac{DSM} with noise. This indicates that adding stochasticity to the \ac{DSM} introduces structural variations that assist the denoiser in the diffusion process. This also demonstrates how diffusion naturally aligns with the ground filtering task, treating non-terrain elements as noise to be systematically removed.

\section{Additional Qualitative Results}
\label{sec:supp_qualitative}

\subsection{DTM Generation}

We provide qualitative results of \ac{method}'s performance across diverse environments and challenging scenarios.

\begin{figure*}[ht]
\centering
\setlength{\tabcolsep}{1pt}     
\renewcommand{\arraystretch}{0.5}
\resizebox{\textwidth}{!}{%
\begin{tabular}{@{} >{\centering\arraybackslash}m{0.55cm} *{6}{>{\centering\arraybackslash}m{0.157\linewidth}} @{}}
& \multicolumn{4}{c}{\textbf{USGS}} & \multicolumn{2}{c}{\textbf{ALS2DTM}} \\
\cmidrule(lr){2-5}\cmidrule(lr){6-7}
& {\small SU-II~\cite{su_2020}} 
& {\small SU-III~\cite{su_2020}} 
& {\small RT~\cite{rt_2018}} 
& {\small KW~\cite{kw_2020}} 
& {\small DALES~\cite{le2022learning}} 
& {\small NB~\cite{le2022learning}} \\
\cmidrule(lr){2-5}\cmidrule(lr){6-7}
\rotatebox{90}{\small Sat. Image} &
\img{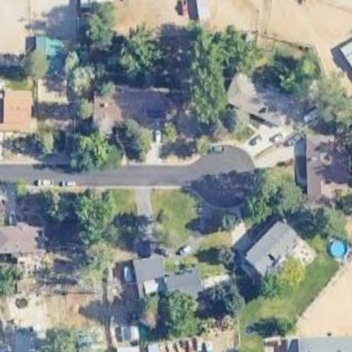} &
\img{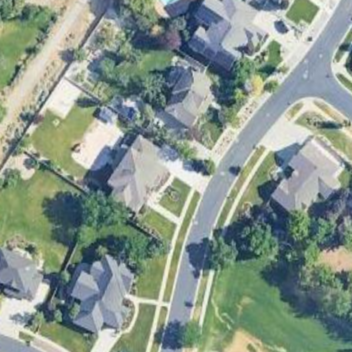} &
\img{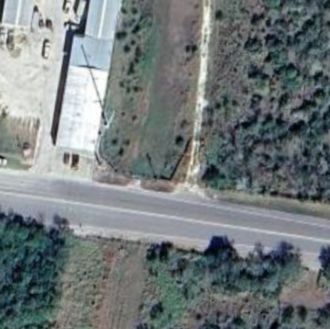} &
\img{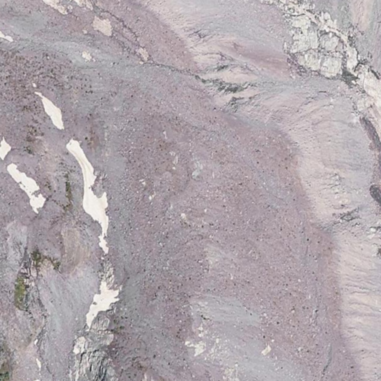} &
\img{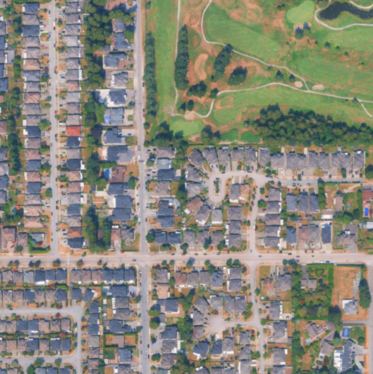} &
\img{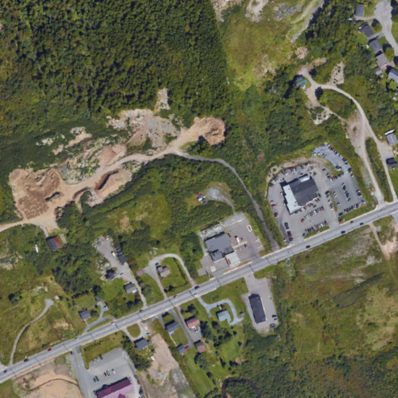} \\
\rotatebox{90}{\small Ground Prob.} &
\img{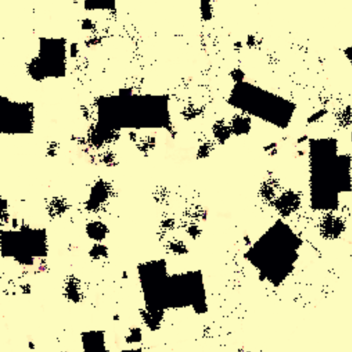} &
\img{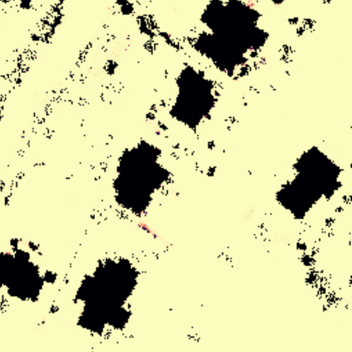} &
\img{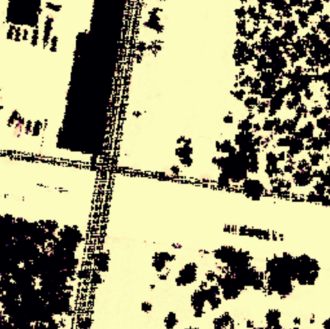} &
\img{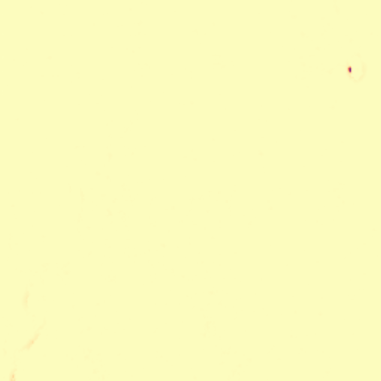} &
\img{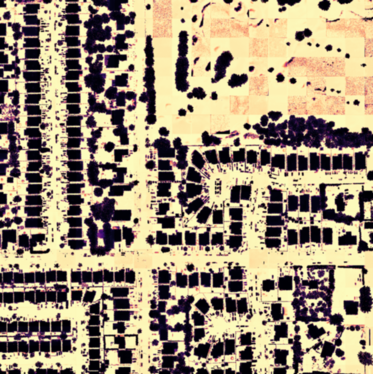} &
\img{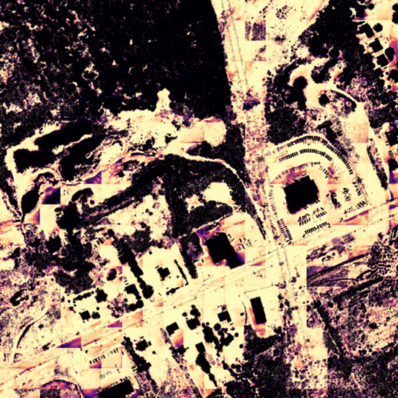} \\
 &
 \barelement{figures/dtm_generation_results/probs_bar.png}{0.0}{1.0} &
 \barelement{figures/dtm_generation_results/probs_bar.png}{0.0}{1.0} &
 \barelement{figures/dtm_generation_results/probs_bar.png}{0.0}{1.0} &
 \barelement{figures/dtm_generation_results/probs_bar.png}{0.0}{1.0} &
 \barelement{figures/dtm_generation_results/probs_bar.png}{0.0}{1.0} &
 \barelement{figures/dtm_generation_results/probs_bar.png}{0.0}{1.0} \\
\rotatebox{90}{\small DSM} &
\img{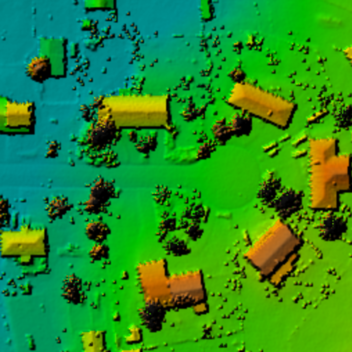} &
\img{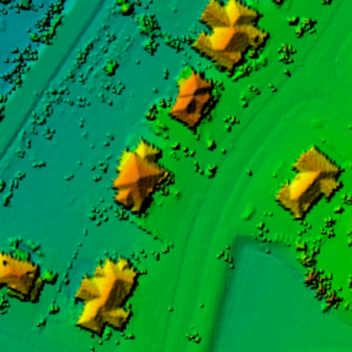} &
\img{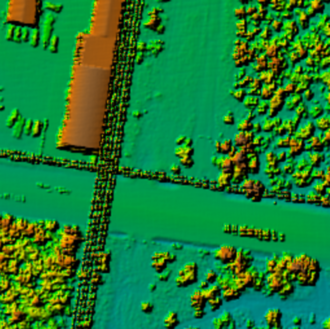} &
\img{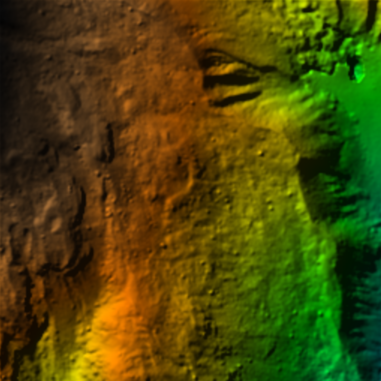} &
\img{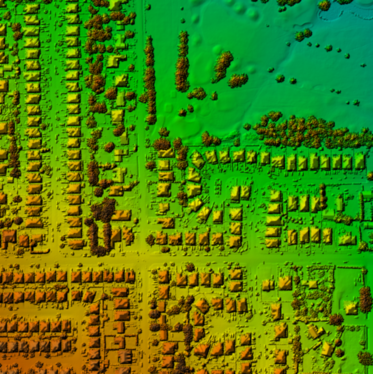} &
\img{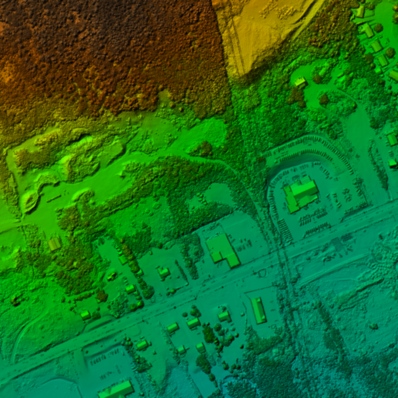} \\
\rotatebox{90}{\small GT DTM} &
\img{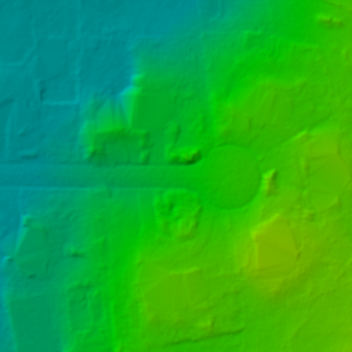} &
\img{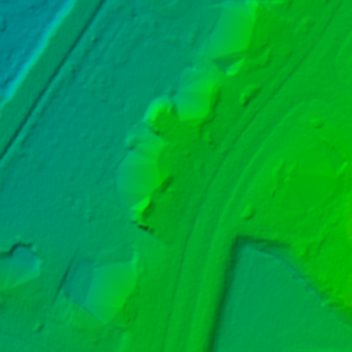} &
\img{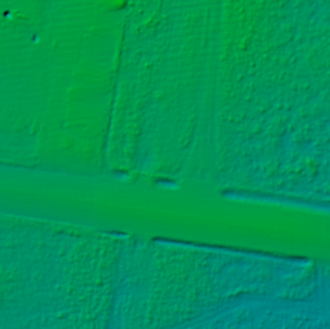} &
\img{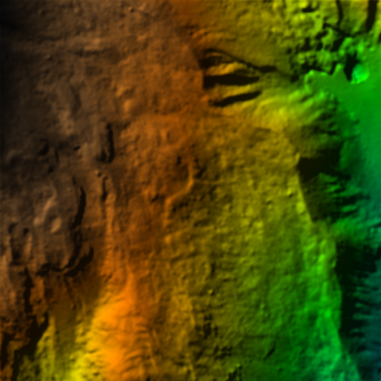} &
\img{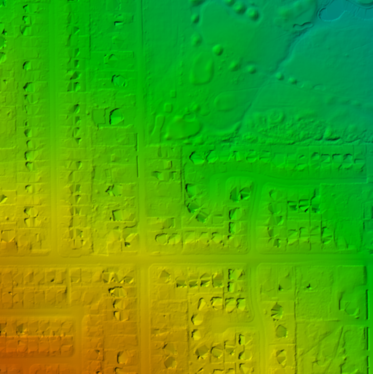} &
\img{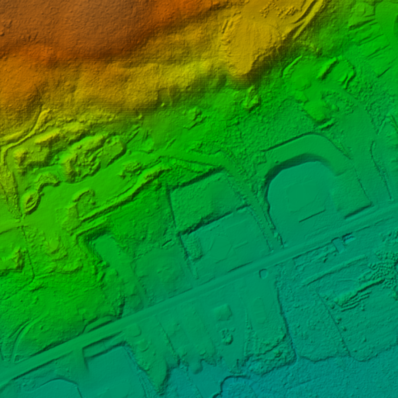} \\
\rotatebox{90}{\small Pred.~DTM} &
\img{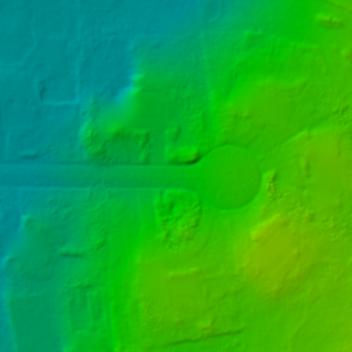} &
\img{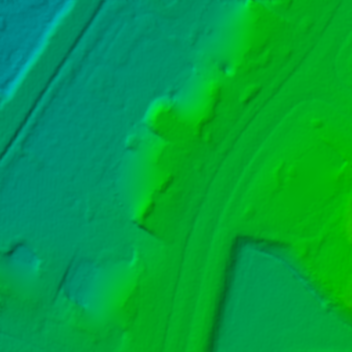} &
\img{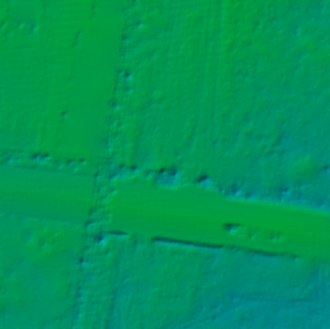} &
\img{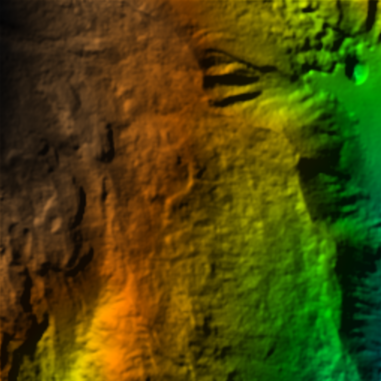} &
\img{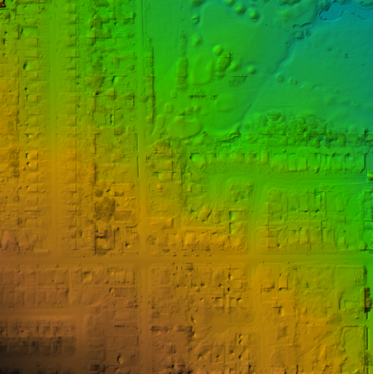} &
\img{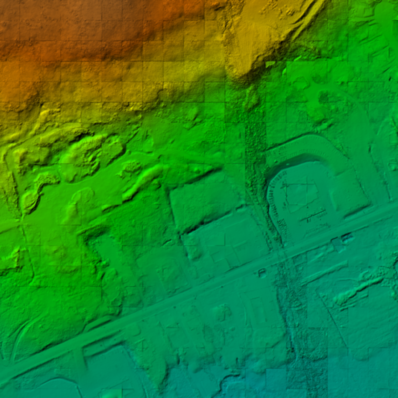} \\
 &
 \barelement{figures/dtm_generation_results/elev_bar.png}{1445m}{1460m} &
 \barelement{figures/dtm_generation_results/elev_bar.png}{1390m}{1410m} &
 \barelement{figures/dtm_generation_results/elev_bar.png}{14.5m}{23m} &
 \barelement{figures/dtm_generation_results/elev_bar.png}{1680m}{1850m} &
 \barelement{figures/dtm_generation_results/elev_bar.png}{12m}{65m} &
 \barelement{figures/dtm_generation_results/elev_bar.png}{40m}{120m} \\
\rotatebox{90}{\small Rel.~error} &
\img{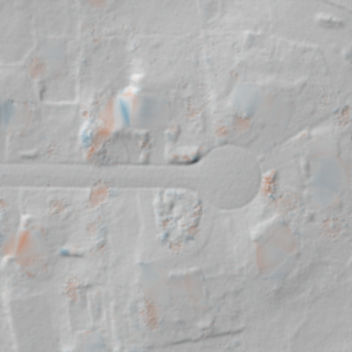} &
\img{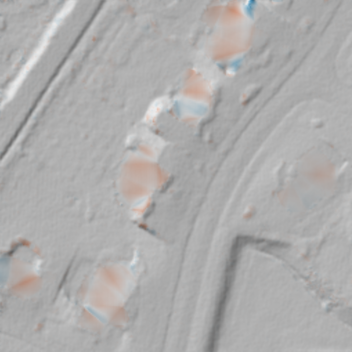} &
\img{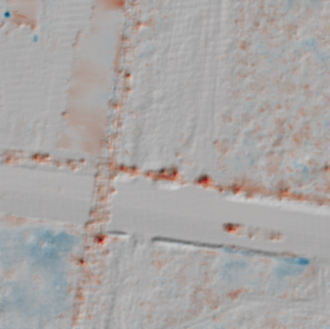} &
\img{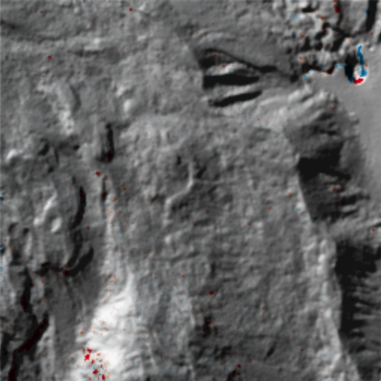} &
\img{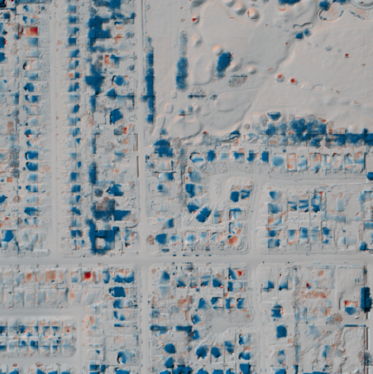} &
\img{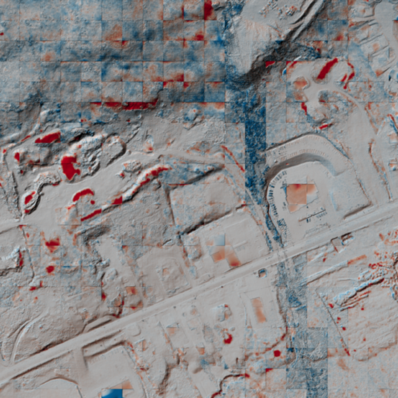} \\
&
\barelement{figures/dtm_generation_results/error_bar.png}{-1.0m}{1.0m}&
\barelement{figures/dtm_generation_results/error_bar.png}{-1.0m}{1.0m}&
\barelement{figures/dtm_generation_results/error_bar.png}{-1.0m}{1.0m}&
\barelement{figures/dtm_generation_results/error_bar.png}{-1.0m}{1.0m}&
\barelement{figures/dtm_generation_results/error_bar.png}{-1.0m}{1.0m}&
\barelement{figures/dtm_generation_results/error_bar.png}{-1.0m}{1.0m} \\
\end{tabular}
} 
\caption{\textbf{Ground generation results.} From top to bottom: satellite imagery, ground probability map, input \ac{DSM}, ground-truth \ac{DTM}, predicted \ac{DTM}, and relative error. Examples cover diverse environments: urban regions (SU-II, SU-III~\cite{su_2020}), suburban areas (RT~\cite{rt_2018}, NB~\cite{le2022learning}), steep mountainous terrain with gentle elevation changes (KW~\cite{kw_2020}), and urban areas (DALES~\cite{le2022learning}). Satellite imagery is from Google Maps and may not be temporally aligned with the geospatial data due to differences in capture dates.}
\label{fig:supp_comprehensive_results}
\vspace{-1\baselineskip}
\end{figure*}

\cref{fig:supp_comprehensive_results} presents a comprehensive overview of our method's performance across all six test datasets. The ground probability maps demonstrate how our model confidently identifies terrain versus above-ground structures, with bright regions indicating high confidence in ground classification. The error maps reveal that most inaccuracies occur beneath buildings and in densely vegetated areas, where true ground measurements are unavailable. In these regions, the ground-truth is typically filled using triangulation-based interpolation. However, our \ac{method} produces physically plausible surface reconstructions that show higher errors, while still better reflecting the actual scene. Importantly, in regions densely covered with vegetation where the ground is nearly invisible (e.g., RT dataset~\cite{rt_2018}), our method still produces reasonable surface predictions.

These additional results further demonstrate \ac{method}'s robustness across diverse environments and its ability to handle challenging scenarios with reasonable performance.

\subsection{Road Reconstruction}
\begin{figure*}[ht]
\centering
\setlength{\tabcolsep}{1pt} 
\renewcommand{\arraystretch}{0.8}
\resizebox{\textwidth}{!}{%
\begin{tabular}{ccccc}
{\ac{DSM}} & {GT \ac{DTM}} & {FlexRoad~\cite{flexroad}} & {\ac{method}} & {\ac{method}+} \\[0.2em]

\includegraphics[width=0.2\linewidth]{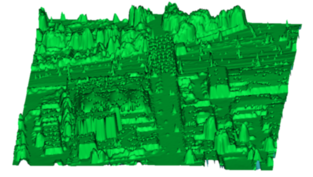} &
\includegraphics[width=0.2\linewidth]{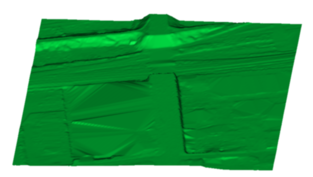} &
\includegraphics[width=0.2\linewidth]{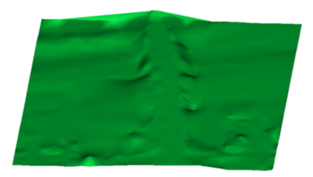} &
\includegraphics[width=0.2\linewidth]{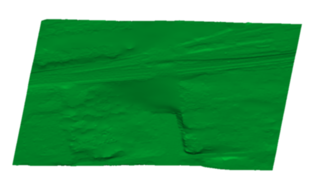} &
\includegraphics[width=0.2\linewidth]{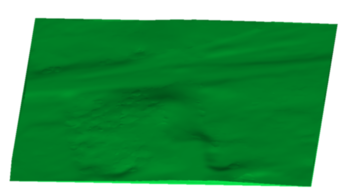} \\[0.2em]

\includegraphics[width=0.2\linewidth]{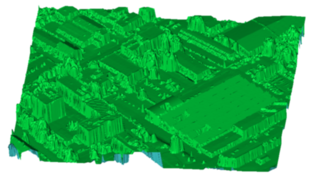} &
\includegraphics[width=0.2\linewidth]{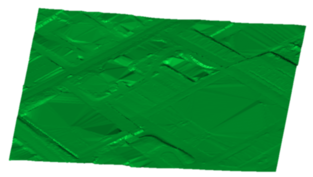} &
\includegraphics[width=0.2\linewidth]{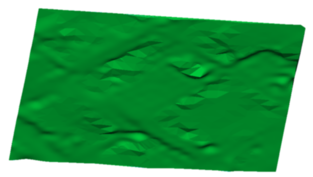} &
\includegraphics[width=0.2\linewidth]{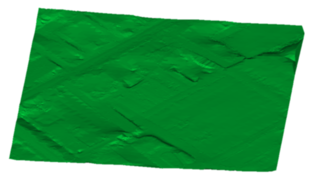} &
\includegraphics[width=0.2\linewidth]{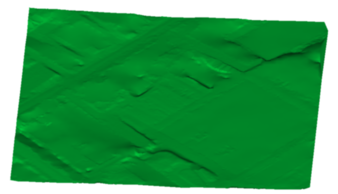} \\[0.2em]

\includegraphics[width=0.2\linewidth]{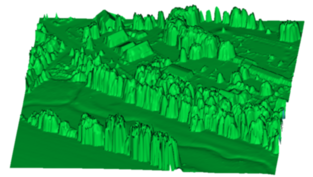} &
\includegraphics[width=0.2\linewidth]{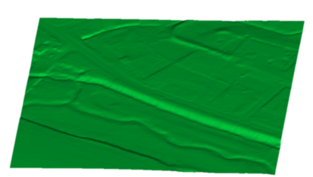} &
\includegraphics[width=0.2\linewidth]{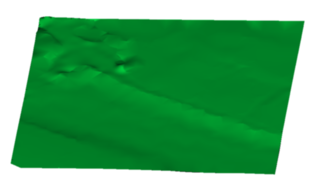} &
\includegraphics[width=0.2\linewidth]{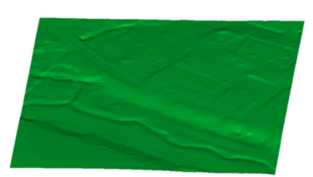} &
\includegraphics[width=0.2\linewidth]{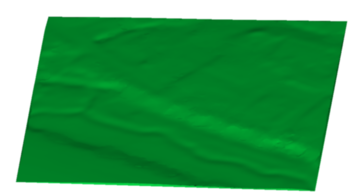} \\[0.2em]

\includegraphics[width=0.2\linewidth]{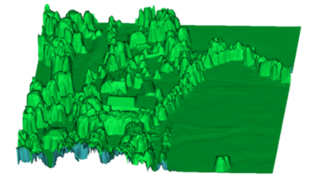} &
\includegraphics[width=0.2\linewidth]{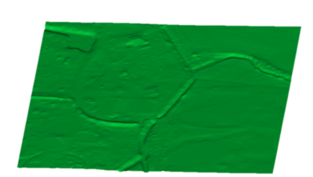} &
\includegraphics[width=0.2\linewidth]{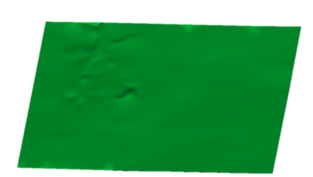} &
\includegraphics[width=0.2\linewidth]{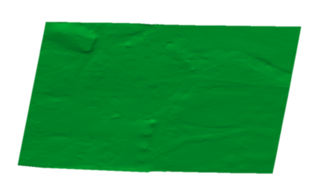} &
\includegraphics[width=0.2\linewidth]{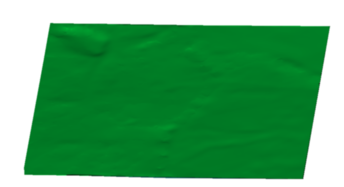} \\[0.2em]

\includegraphics[width=0.2\linewidth]{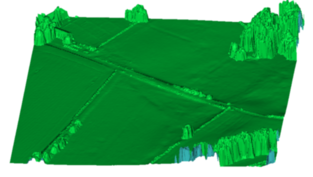} &
\includegraphics[width=0.2\linewidth]{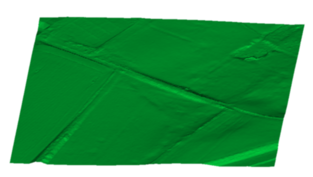} &
\includegraphics[width=0.2\linewidth]{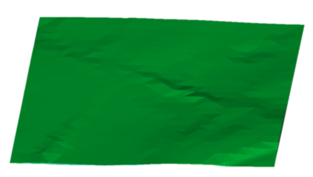} &
\includegraphics[width=0.2\linewidth]{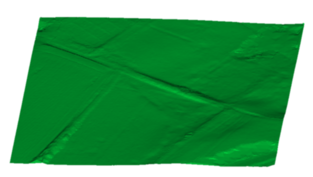} &
\includegraphics[width=0.2\linewidth]{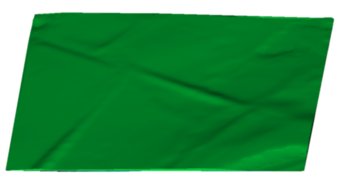} \\

\end{tabular}
}
\caption{\textbf{3D mesh visualizations for road reconstruction from samples in the GeRoD dataset~\cite{flexroad}.} Each row shows a different scene comparing the input \ac{DSM} (left), ground-truth \ac{DTM}, FlexRoad~\cite{flexroad}, our \ac{method}, and its smoothness-enhanced version \ac{method}+ (right). Our method recovers the underlying terrain more accurately, while \ac{method}+ achieves higher smoothness while maintaining high precision.}
\label{fig:road_mesh_comparison}
\vspace*{-1\baselineskip}
\end{figure*}

\cref{fig:road_mesh_comparison} provides visual comparison of road surface reconstruction across different scenarios. For urban regions with bridges (first row), FlexRoad~\cite{flexroad} can model elevated bridge structures because it uses segmentation-based road extraction and fits NURBS surfaces to identified road segments. In contrast, our \ac{method} is trained to remove all above-ground structures including bridges, making it more accurate at modeling the underlying terrain and tunnel areas beneath bridges while sacrificing elevated road surface representation. Additionally, classification artifacts from ground detection may result in incomplete modeling of road surfaces on bridges. By training our model on data where bridges are part of the \ac{DTM}, road modeling could be completely handled by our method.

Across all scenes, \ac{method} produces visually more coherent surfaces with structurally plausible terrain continuity. In all cases demonstrates superior road edge modeling compared to FlexRoad~\cite{flexroad}, with cleaner transitions between road surfaces and adjacent terrain. Our method effectively handles abrupt elevation variations and discontinuities in road surfaces, producing more accurate local topography.

However, the fine-grained mesh details in our reconstructions, while geometrically accurate, result in slightly reduced surface smoothness compared to FlexRoad's mathematically constrained NURBS approach. Our extended version, \ac{method}+, further improves smoothness while remaining flexible, preserving sharp transitions and fine details without significantly compromising precision.

\section{Ablations on GrounDiff}
\label{sec:supp_groundiff}

We report comprehensive ablation results on the GeRoD dataset~\cite{flexroad}, including reverse diffusion initialization schemes, loss functions, and normalization strategies. The dataset splits follow the description in the main paper.

\begin{table}[ht]
\centering
\resizebox{\linewidth}{!}{%
\setlength{\tabcolsep}{2pt}
\footnotesize
\begin{tabular}{lccccc}
\toprule
\bfseries{Variant} & \bfseries{RMSE}$\downarrow$ & \bfseries{MAE}$\downarrow$ & \bfseries{$\boldsymbol{E_{T_1}}$}$\downarrow$ & \bfseries{$\boldsymbol{E_{T_2}}$}$\downarrow$ & \bfseries{$\boldsymbol{E_{tot}}$}$\downarrow$ \\
\midrule
Init: Noise & 0.723 & 0.401 & 1.43 & 1.06 & 1.11 \\
Init: DSM & 0.715 & 0.400 & 1.45 & 1.05 & 1.13 \\
Loss: $\mathcal{L}_1$ & 0.742 & 0.412 & 1.56 & \textbf{0.68} & \underline{1.01} \\
Loss: $\mathcal{L}_1$ + $\mathcal{L}_2$ & \underline{0.708} & \textbf{0.383} & \underline{1.31} & \underline{0.74} & \textbf{0.93} \\
Loss: $\mathcal{L}_1$ + $\mathcal{L}_2$ + $\mathcal{L}_{\nabla}$ & 0.825 & 0.439 & 1.54 & 0.89 & 1.11 \\
Norm: Data Localization~\cite{amirkolaee2022dtm} & 7.279 & 4.692 & 16.42 & 17.69 & 15.80 \\
Norm: Global Standardization~\cite{bittner2023dsm2dtm} & 0.950 & 0.556 & \textbf{1.03} & 2.14 & 1.37 \\
\midrule
\textbf{Baseline (Ours)} & \textbf{0.700} & \underline{0.393} & 1.43 & 1.06 & 1.11 \\
\bottomrule
\end{tabular}
}
\caption{\textbf{Extended ablation studies of \ac{method} on GeRoD dataset~\cite{flexroad}.} 
Includes initialization, loss functions, and normalization.}
\label{tab:architecture_ablation_appendix}
\vspace*{-1\baselineskip}
\end{table}

The results in \cref{tab:architecture_ablation_appendix} provide a detailed analysis of each design choice. Initializing the reverse diffusion with either pure noise or the \ac{DSM} yields competitive results, with the \ac{DSM} performing slightly better by providing a structured prior while maintaining stochasticity. Combining stochasticity with the \ac{DSM} further improves performance, supporting the idea of treating the \ac{DSM} structure as a form of noise. Using only the $\mathcal{L}_1$ loss increases both RMSE and MAE, whereas combining $\mathcal{L}_1 + \mathcal{L}_2$ improves the overall trade-off between height accuracy and classification metrics. Including the gradient loss $\mathcal{L}_{\nabla}$ without gating slightly increases errors. Regarding normalization, min-max normalization outperforms both global standardization and data localization, demonstrating the benefit of scale-agnostic learning across varied terrain heights. 

Collectively, these extended ablations, together with those in the main paper, reinforce the design choices of our baseline method and highlight the components essential for robust terrain reconstruction.

\section{Ablations on PrioStitch}
\label{sec:supp_stitching}

\begin{figure*}[htp]
\centering
\setlength{\tabcolsep}{3pt}

\begin{tabular}{ccc}

\includegraphics[width=0.22\linewidth]{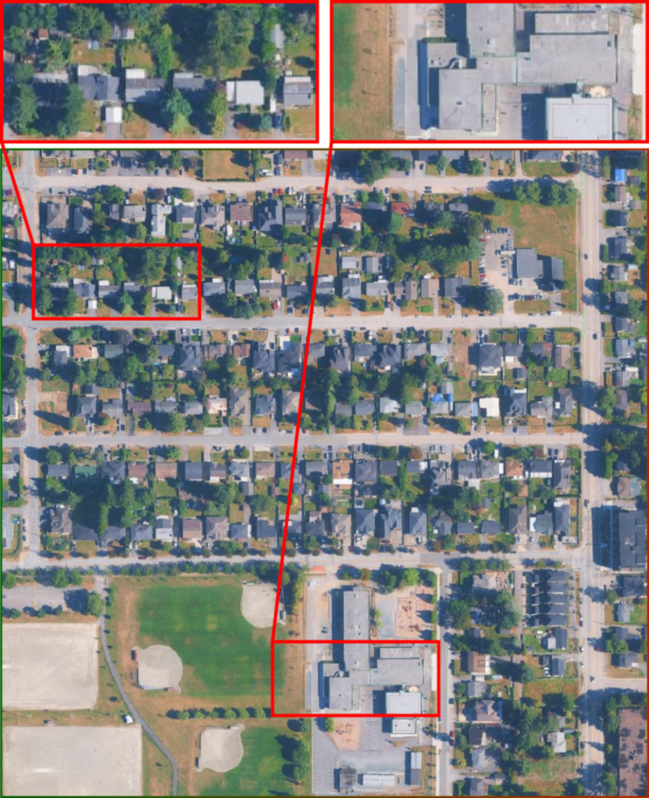} &
\includegraphics[width=0.22\linewidth]{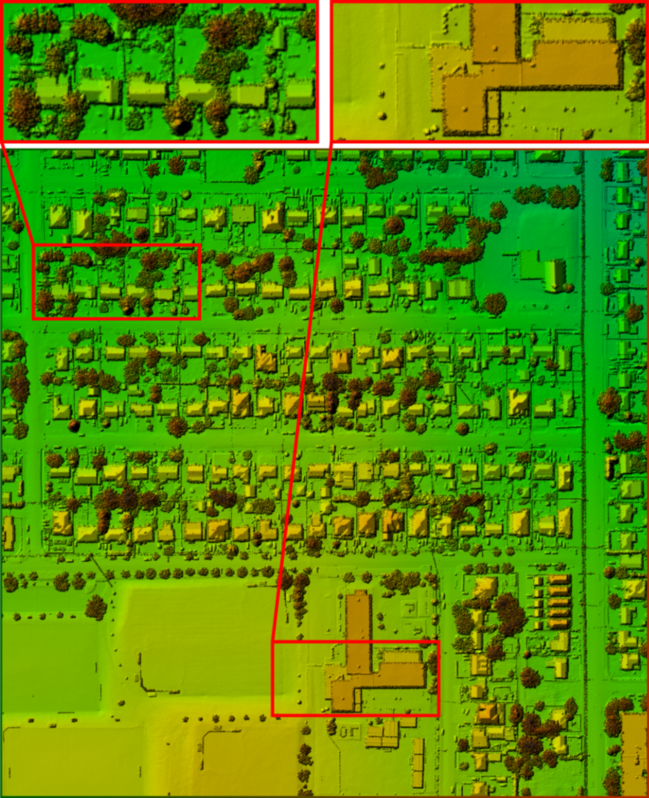} &
\includegraphics[width=0.22\linewidth]{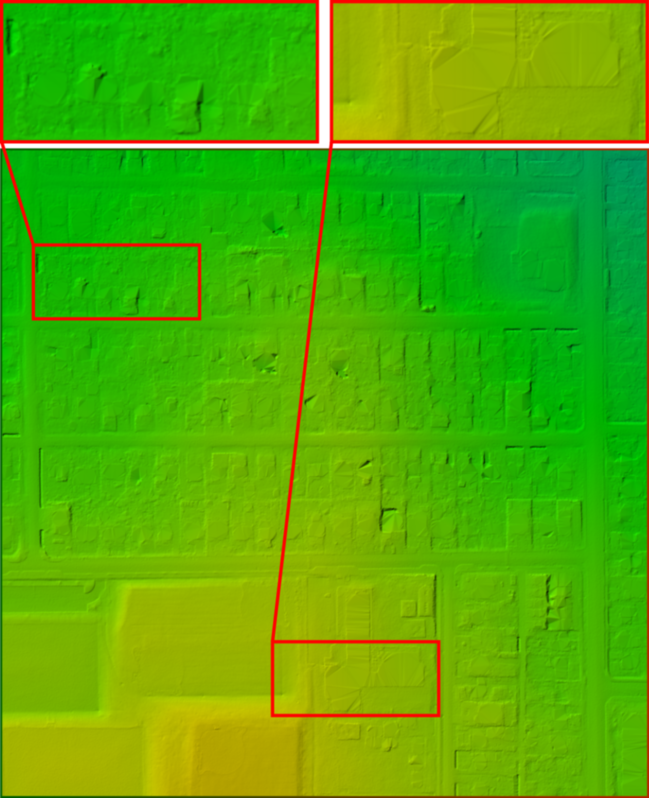} \\
{\scriptsize Sat. Image} & {\scriptsize DSM} & {\scriptsize GT DTM} \\[0.5em]

\includegraphics[width=0.22\linewidth]{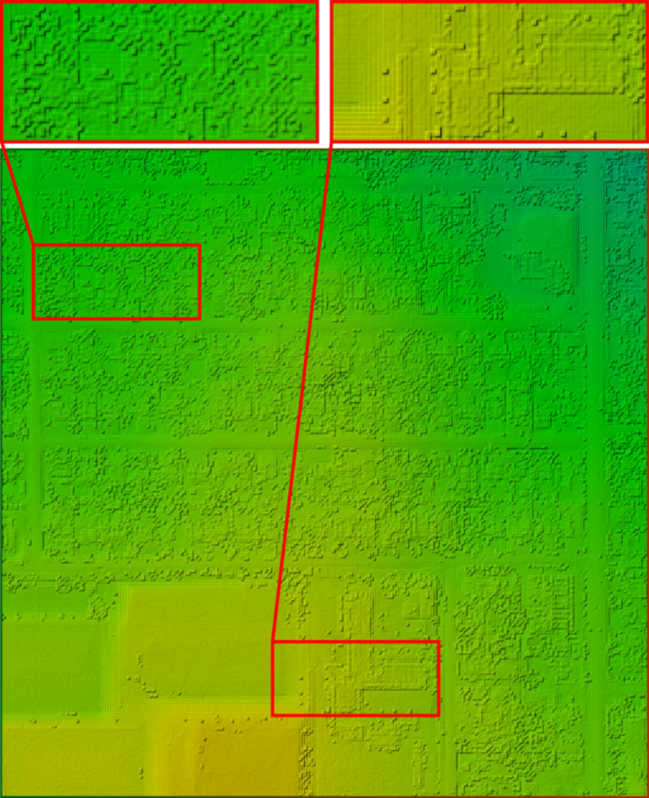} &
\includegraphics[width=0.22\linewidth]{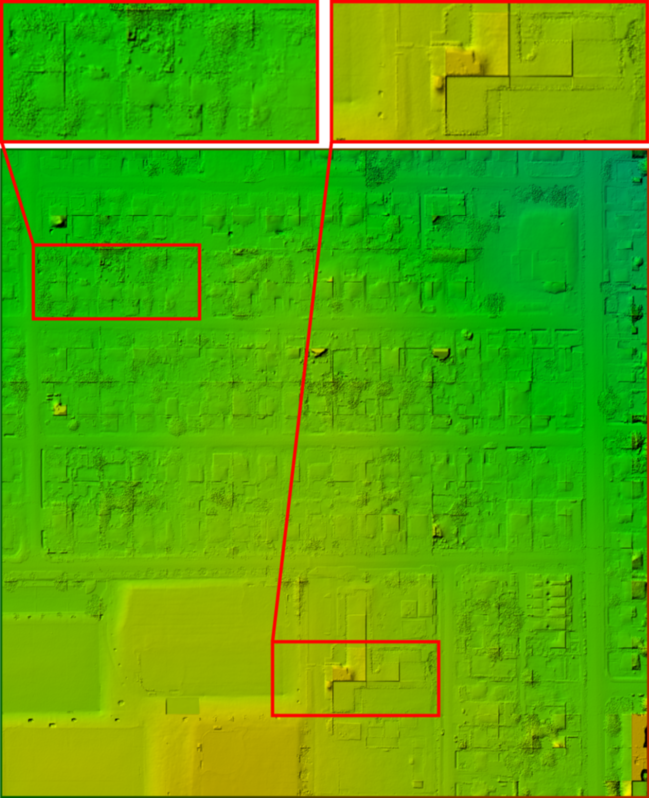} &
\includegraphics[width=0.22\linewidth]{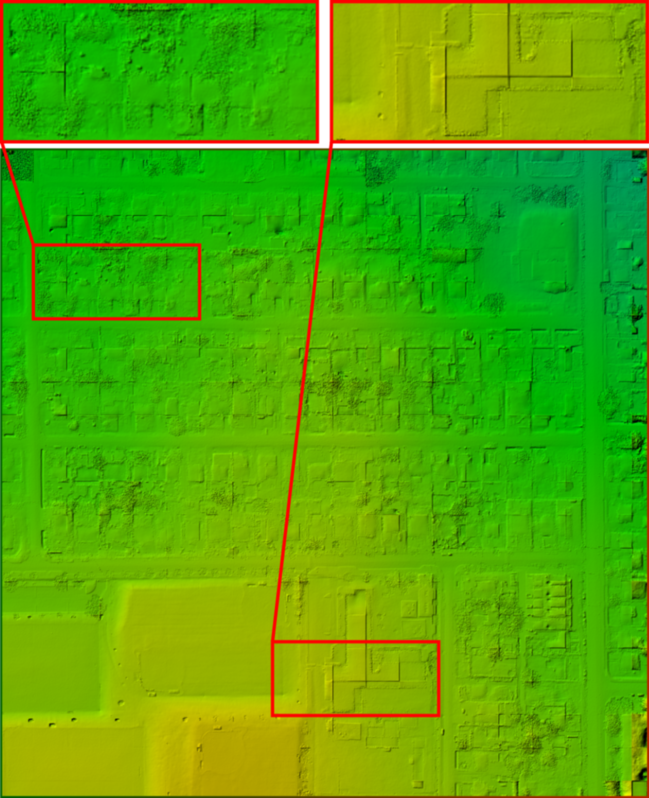} \\
{\scriptsize (a) w/o \ac{stitching}} & {\scriptsize (b) \ac{stitching} w/o prior downsampled \ac{DTM}} & {\scriptsize (c) \ac{stitching} w/o overlapping} \\[0.5em]

\includegraphics[width=0.22\linewidth]{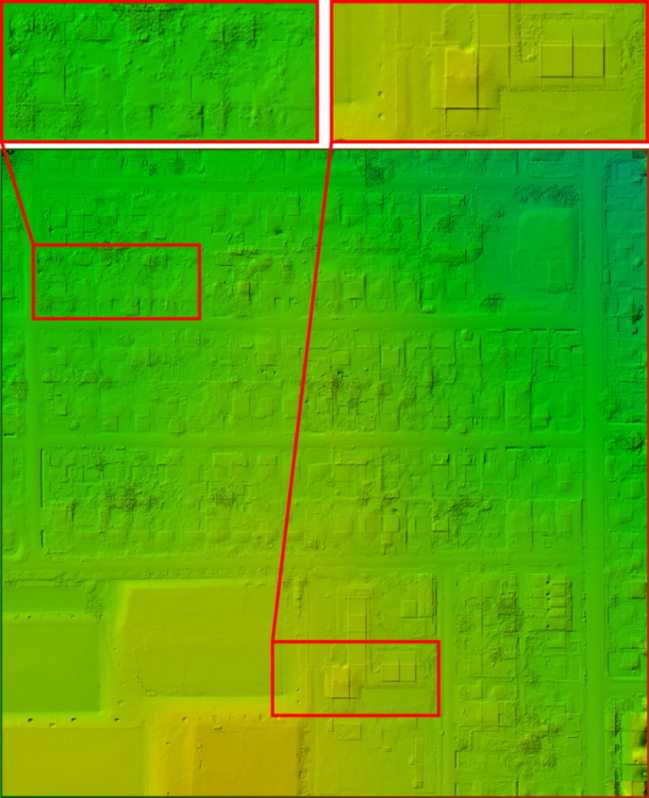} &
\includegraphics[width=0.22\linewidth]{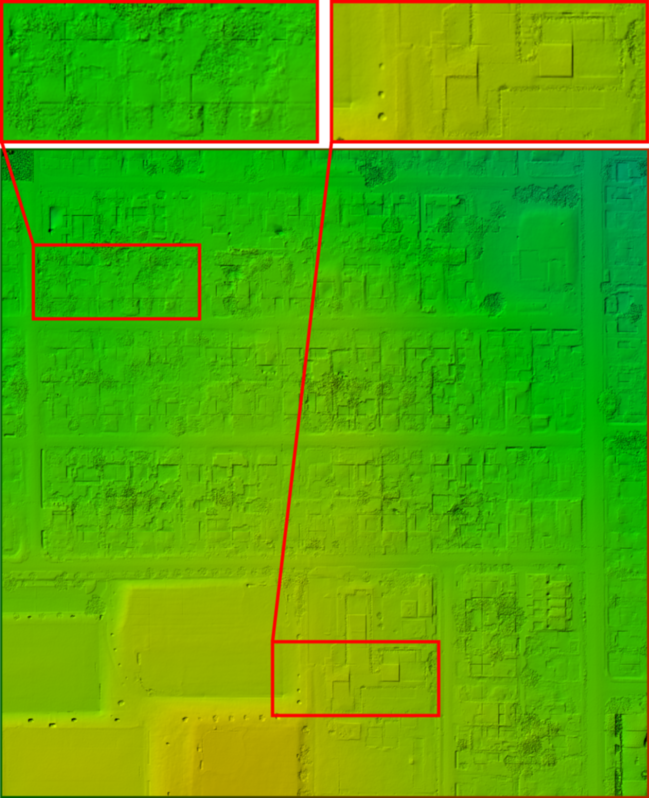} &
\includegraphics[width=0.22\linewidth]{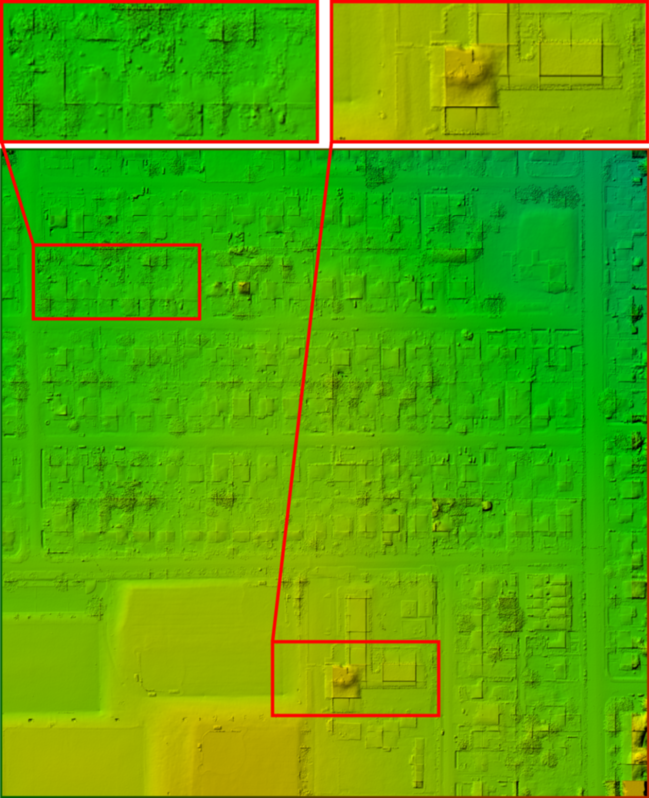} \\
{\scriptsize (d) \ac{stitching} with mean blending} & {\scriptsize (e) \ac{stitching} with min blending} & {\scriptsize (f) \ac{stitching} with max blending} \\[0.5em]

\includegraphics[width=0.22\linewidth]{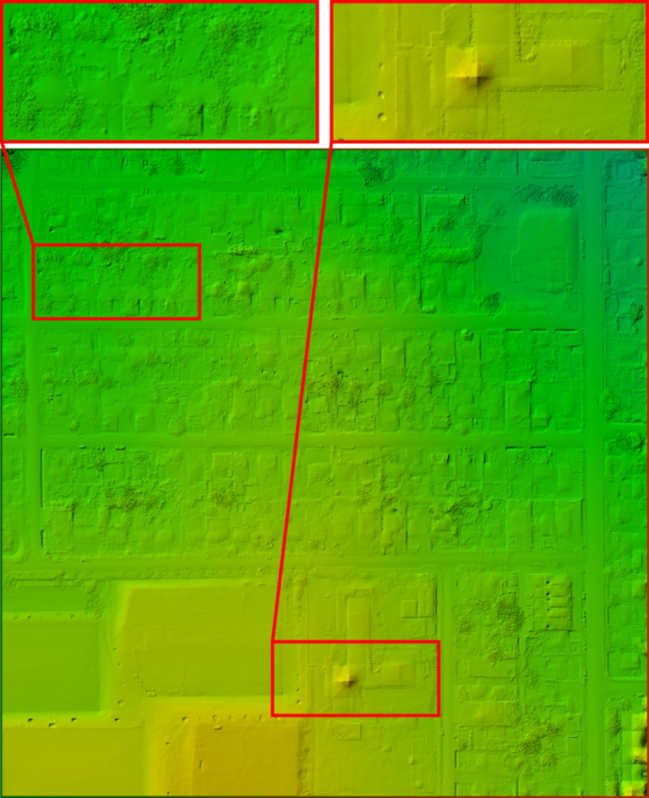} &
\includegraphics[width=0.22\linewidth]{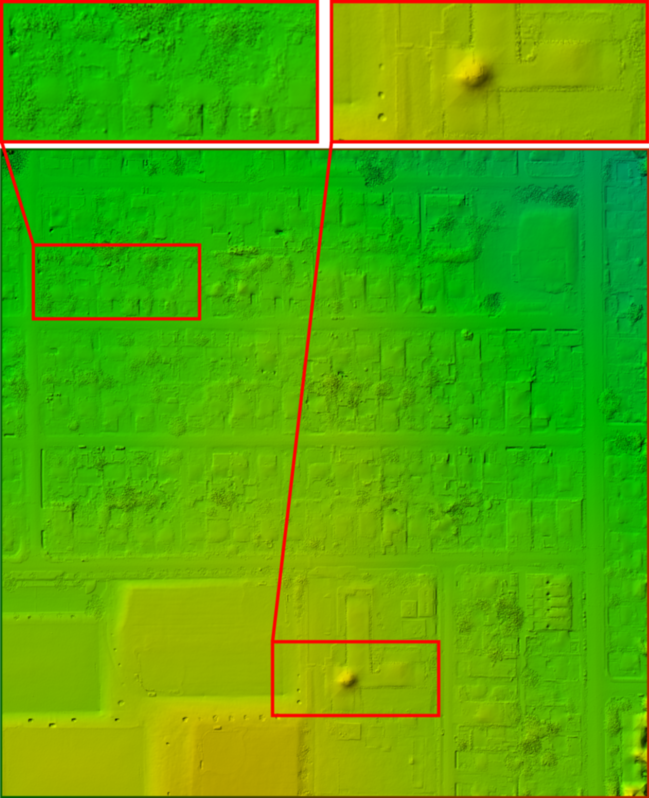} &
\includegraphics[width=0.22\linewidth]{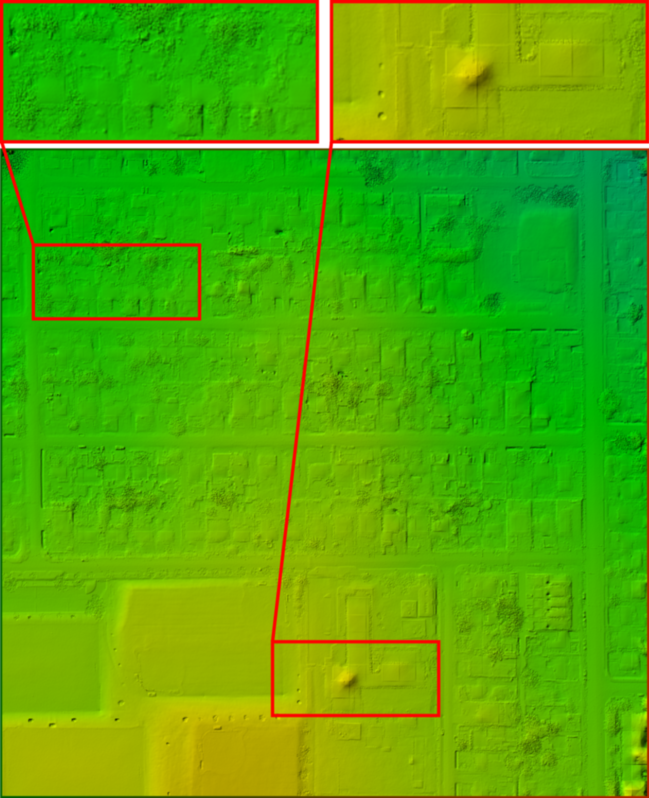} \\
{\scriptsize (g) \ac{stitching} with linear blending} & {\scriptsize (h) \ac{stitching} with cosine blending} & {\scriptsize (i) \ac{stitching} with exponential blending} \\

\end{tabular}

\caption{\textbf{Visual comparison of our \ac{stitching} ablations.} We show a random large-scale test sample (500~m $\times$ 500~m at 0.1~m/pixel) from the DALES~\cite{le2022learning} dataset. The predicted \acp{DTM} (a–i) correspond to our model with the configurations defined in~\cref{tab:priostich_ablation}.}

\label{fig:priostitch_ablation}
\vspace*{-1\baselineskip}
\end{figure*}

We conduct detailed evaluation of our \ac{stitching} approach on the large-scale urban DALES dataset~\cite{le2022learning}. \cref{tab:priostich_ablation} provides quantitative results for different configurations. \cref{fig:priostitch_ablation} provides visual comparison of the different approaches.

\begin{table}[ht]
\centering
\resizebox{\linewidth}{!}{%
\begin{tabular}{lcccccccc}
\toprule
& \bfseries{Prio} & \bfseries{Stitching} & \bfseries{Overlap} & \bfseries{Mode} & \bfseries{RMSE}$\downarrow$ & \bfseries{MAE}$\downarrow$ & \bfseries{$\boldsymbol{E_{tot}}$}$\downarrow$ & \bfseries{MAD}$\downarrow$ \\
\midrule
(a) & \xmark & \xmark & \xmark & - & 0.780 & 0.269 & 17.80 & \textbf{3.35} \\
(b) & \xmark & \cmark & \xmark & - & 0.911 & 0.321 & 9.31 & 12.85\\
(c) & \cmark & \cmark & \xmark & - & 0.708 & 0.256 & 8.40 & 13.07\\
(d) & \cmark & \cmark & \cmark & mean & \underline{0.600} & 0.230 & 7.68 & 12.23\\
(e) & \cmark & \cmark & \cmark & min & \textbf{0.514} & \textbf{0.196} & \textbf{7.63} & 12.86\\
(f) & \cmark & \cmark & \cmark & max & 0.941 & 0.359 & 9.61 & 13.37\\
(g) & \cmark & \cmark & \cmark & linear & 0.608 & \underline{0.224} & 7.68 & \underline{11.87}\\
(h) & \cmark & \cmark & \cmark & cosine & 0.623 & 0.226 & 7.75 & 12.00\\
(i) & \cmark & \cmark & \cmark & exp & 0.605 & \underline{0.224} & \underline{7.67} & 12.00\\
\bottomrule
\end{tabular}%
}
\caption{\textbf{Ablation study of the \ac{stitching} strategy on the DALES dataset~\cite{le2022learning}.} Systematic evaluation of tiling and blending components. \textbf{Prior}: whether a low-resolution \ac{DTM} is used for initialization, otherwise noisy DSM is used; \textbf{Stitching}: whether the input \ac{DSM} is processed in tiles; \textbf{Overlap}: whether tiles overlap by 50 percent, stride 128; \textbf{Mode}: blending strategy for merging overlapping regions. Metrics include RMSE and MAE in meters, total classification error in percent, and MAD in degrees measuring surface roughness. Ground-truth \acp{DTM} have an MAD of 3.33 degrees. \textbf{Bold} indicates best performance, \underline{underlined} indicates second best.}
\label{tab:priostich_ablation}
\vspace*{-1\baselineskip}
\end{table}

\subsection{Impact of Global Prior}

When \ac{stitching} is not applied and no prior data is used (a), the network processes a downscaled \ac{DSM} where downsampling and upsampling introduce interpolation artifacts and eliminate fine-grained details, inducing high regression and classification errors (RMSE=0.780, $E_{tot}$=17.80\%). However, this approach achieves natural smoothness (MAD=3.35°) closest to ground-truth terrain (3.33°) due to the inherent smoothing effect of downsampling that removes small non-ground elements.

Enabling stitching without prior conditioning (b) significantly reduces classification error to 9.31\% because the network can observe fine details in full-resolution tiles. However, RMSE increases and surface roughness worsens (MAD = 12.85\textdegree) due to limited contextual information: some tiles contain only vegetation or buildings without visible ground, which challenges accurate terrain prediction.

Incorporating a low-resolution prior \ac{DTM} auto-generated using \ac{method} (as in configurations (a)) (c) provides essential global context, reducing regression errors (RMSE = 0.708) by 22\,\% and classification errors ($E_\text{tot}$ = 8.40\,\%) by 10\,\% compared to configuration (b). The prior guides consistent terrain interpretation across ambiguous regions, although surface roughness remains elevated (MAD = 13.07\textdegree) compared to the naturally smooth downsampled approach.

\subsection{Blending Strategies}
\begin{figure}[ht]
    \centering
    \includegraphics[width=\linewidth]{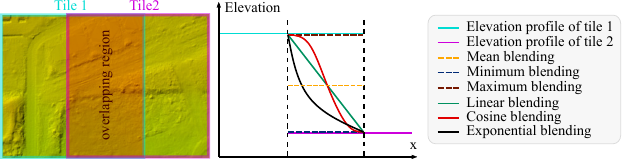}
    \caption{\textbf{Blending strategy weighting functions.} Visualization of two overlapping tiles with constant elevation profiles (for simplicity) showing inconsistencies in the overlap region. Different blending modes are applied to merge predictions, demonstrating how each weighting strategy fuses elevation data and handles boundary discontinuities in our \ac{stitching} approach.}
    \label{fig:modes}
    \vspace*{-1\baselineskip}
\end{figure}

We evaluate several blending strategies for merging overlapping tile outputs, as shown in~\cref{fig:modes}:
\begin{itemize}
    \item \textbf{Mean}: Simple averaging of overlapping regions.
    \item \textbf{Min}: Taking the minimum elevation at each overlap point.
    \item \textbf{Max}: Taking the maximum elevation at each overlap point.
    \item \textbf{Linear}: Linear weighting based on distance from tile edge.
    \item \textbf{Cosine}: Cosine-based weighting for smoother transitions.
    \item \textbf{Exponential}: Exponential decay weighting.
\end{itemize}

Using overlapping tiles (d-i) further improves all metrics by increasing ground visibility: when individual tiles contain only non-ground regions (vegetation, buildings), overlapping provides additional spatial context where neighboring tiles are more likely to observe ground surfaces. 

Minimum blending (e) achieves the best performance across all regression and classification metrics, reducing RMSE and MAE by 14\% compared to mean blending (d). This strategy effectively removes residual above-ground artifacts from neighboring tile predictions, as it favors lower elevations that are more likely to represent true ground surfaces. The resulting \ac{DTM} appears closest to ground-truth visually, though it may introduce sharp elevation jumps at tile boundaries. Conversely, maximum blending (f) performs worst as it preserves above-ground artifacts from overlapping predictions. Continuous blending strategies (linear, cosine, exponential) provide smoother boundary transitions, with linear blending (g) offering the best balance of performance and visual quality.

Despite these improvements, the overall MAD remains significantly higher than ground-truth (11.87°-13.07° vs. 3.33°), indicating that some tiles with severely limited ground visibility still produce suboptimal predictions. While prior \ac{DTM} conditioning provides strong guidance toward correct terrain interpretation, the limited input field of view in challenging scenarios prevents perfect reconstruction of natural surface smoothness.

We encourage further research in this direction by exploring high-resolution \ac{DTM} generation with networks supporting arbitrary input sizes, as well as point-based diffusion networks to capture more contextual and global information.

\begin{figure}[ht]
\centering
\setlength{\tabcolsep}{1pt}     
\renewcommand{\arraystretch}{0.5}
\resizebox{\linewidth}{!}{%
\begin{tabular}{@{} >{\centering\arraybackslash}m{0.55cm} *{3}{>{\centering\arraybackslash}m{0.3\linewidth}} @{}}
& \multicolumn{3}{c}{\textbf{Failure Cases}} \\
\cmidrule(lr){2-4}
& {\small Rocky Mountain} 
& {\small Forest} 
& {\small River Crossing Forest} \\
\cmidrule(lr){2-4}
\rotatebox{90}{\small Sat. Image} &
\img{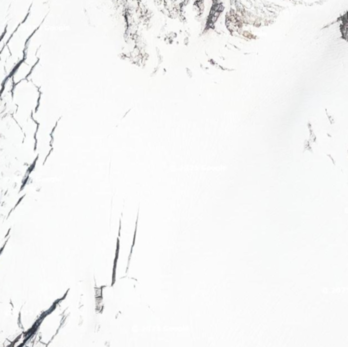} &
\img{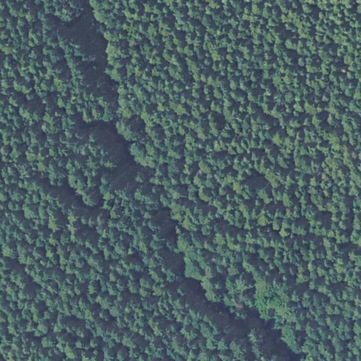} &
\img{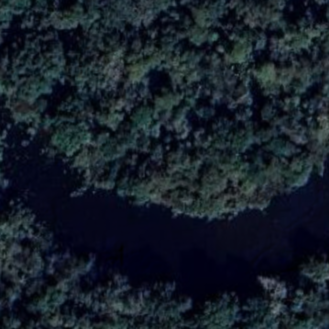} \\
\rotatebox{90}{\small Ground Prob.} &
\img{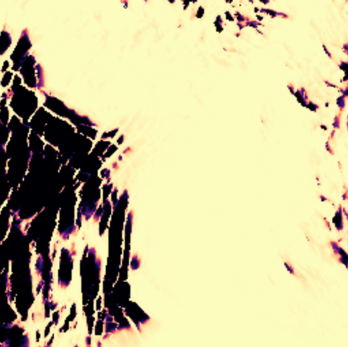} &
\img{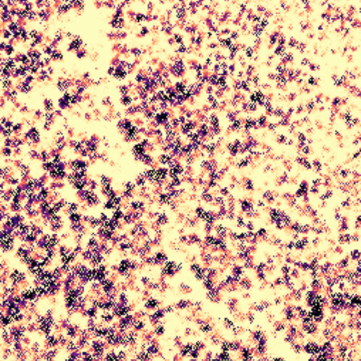} &
\img{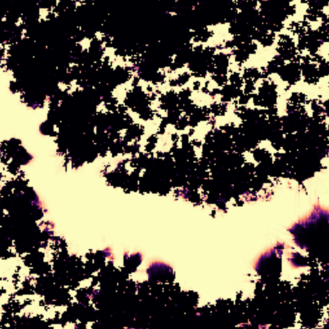} \\
& 
\barelement{figures/dtm_generation_results/probs_bar.png}{0.0}{1.0} &
\barelement{figures/dtm_generation_results/probs_bar.png}{0.0}{1.0} &
\barelement{figures/dtm_generation_results/probs_bar.png}{0.0}{1.0} \\
\rotatebox{90}{\small DSM} &
\img{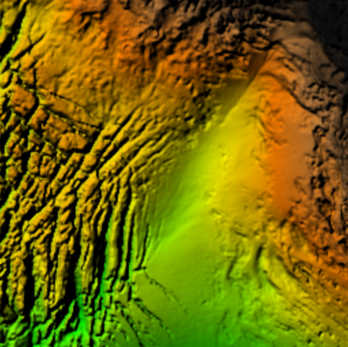} &
\img{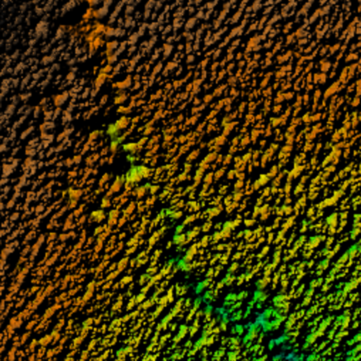} &
\img{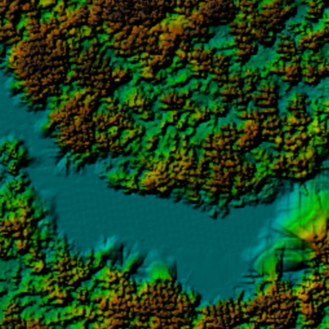} \\
\rotatebox{90}{\small GT DTM} &
\img{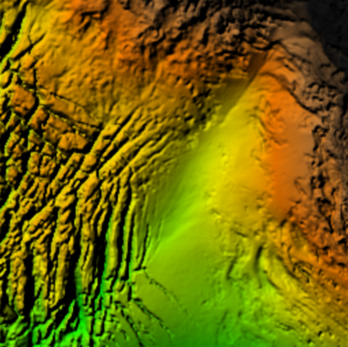} &
\img{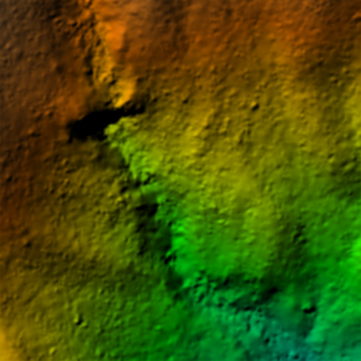} &
\img{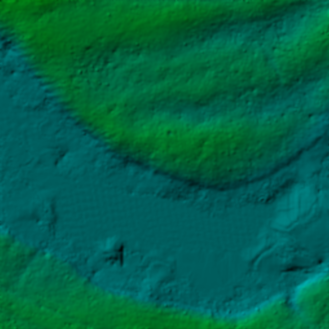} \\
\rotatebox{90}{\small Pred.~DTM} &
\img{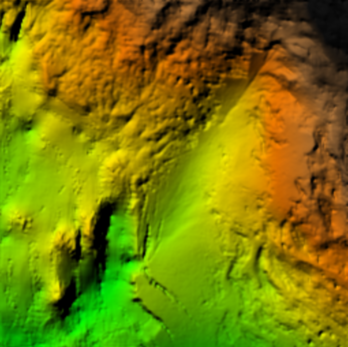} &
\img{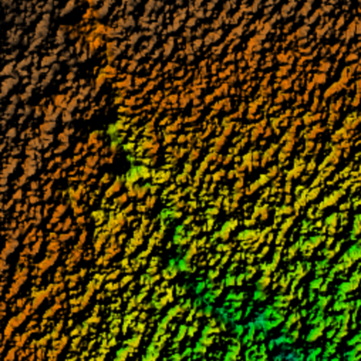} &
\img{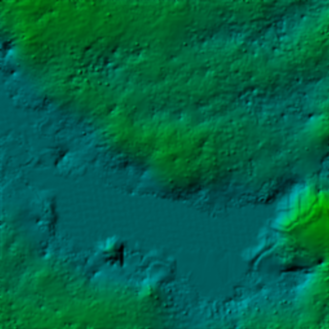} \\
& 
\barelement{figures/dtm_generation_results/elev_bar.png}{2200m}{2300m} &
\barelement{figures/dtm_generation_results/elev_bar.png}{1390m}{1410m} &
\barelement{figures/dtm_generation_results/elev_bar.png}{2m}{20m} \\
\rotatebox{90}{\small Rel.~Error} &
\img{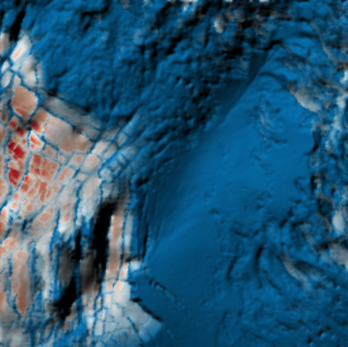} &
\img{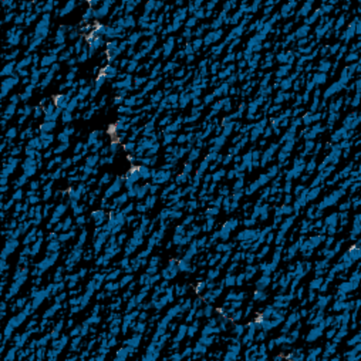} &
\img{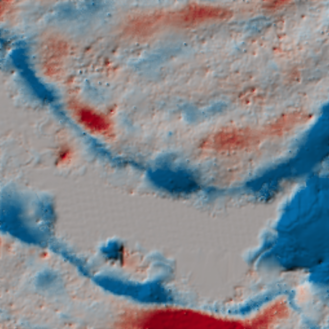} \\
&
\barelement{figures/dtm_generation_results/error_bar.png}{-1.0m}{1.0m}&
\barelement{figures/dtm_generation_results/error_bar.png}{-1.0m}{1.0m}&
\barelement{figures/dtm_generation_results/error_bar.png}{-1.0m}{1.0m}\\
\end{tabular}
} 
\caption{\textbf{Failure case examples for ground generation.} From top to bottom: satellite imagery, ground probability map, input \ac{DSM}, ground-truth \ac{DTM}, predicted \ac{DTM}, and relative error. These examples highlight challenging environments: mountainous regions with abrupt elevation jumps, forested areas, and forested regions with rivers. Satellite imagery is from Google Maps and may not be temporally aligned with the geospatial data due to differences in capture dates.}
\label{fig:failure_cases}
\vspace*{-1\baselineskip}
\end{figure}

\section{Limitations}
\label{sec:supp_limitations}
We show examples of failure cases in~\cref{fig:failure_cases}. All predictions of our \ac{method} are obtained using the model trained on SU-I, where the portion of mountainous and forested regions is very small compared to the overall area, which is predominantly urban. Despite strong performance in urban and suburban regions, our \ac{method} struggles in areas with abrupt elevation changes (e.g., alpine terrain), where sharp elevation gradients resemble those of building facades and are consequently misclassified as non-ground structures, leading to regeneration errors and locally smoothed surfaces. In dense vegetation regions where ground is largely occluded, the model learns to estimate vegetation height but lacks ground reference points in the input data, causing the network to fail completely when elevation differences between pixels are insufficient to identify above-ground structures. Limited ground visibility can also cause the network to hallucinate terrain. However, when a reasonable number of ground pixels are visible, such as along a river crossing a forest, the surface generation becomes more accurate. Future work could leverage \ac{DOP} to enrich semantic features or integrate cross-attention within the encoder–decoder architecture. Even in these challenging areas, the generated regions remain visually and physically plausible, and above-ground structures are typically removed successfully.

\section{Ethical Considerations}
\label{sec:supp_ethic}

Our approach operates exclusively on elevation data, which contains no personally identifiable information or sensitive geographic metadata. The network processes normalized height values without absolute coordinates, ensuring spatial anonymity. All datasets are publicly available, and the ground sampling distance prevents individual identification.

Training data covers diverse regions; however, performance may degrade in environments substantially different from the training distribution. This limitation is most relevant for extreme topographies underrepresented in current datasets, potentially introducing domain-specific biases.

Our diffusion-based approach is probabilistic and cannot provide deterministic accuracy guarantees. The generative nature of the model may introduce reconstruction artifacts, particularly in occluded regions with limited ground visibility. While extensive validation demonstrates robust performance across benchmarks, we recommend domain-specific evaluation before deployment in safety-critical applications requiring high-precision terrain modeling.

The research scope and data characteristics present no ethical considerations beyond standard machine learning best practices.

\end{document}